\documentclass{article}
 \usepackage[final,nonatbib]{nips_2017}

\usepackage[utf8]{inputenc} % allow utf-8 input
\usepackage[T1]{fontenc}    % use 8-bit T1 fonts

\usepackage{caption}
\usepackage{subcaption}
\usepackage{tikz}
\usepackage{xcolor}
\usetikzlibrary{arrows}
\usepackage{threeparttable}
\usepackage{multirow}
\usepackage{cancel}
\usepackage{tabularx}

\usepackage{paralist}
\usepackage{todonotes}

\usepackage{times}
\usepackage{graphicx}
\usepackage{algorithm}
\usepackage{algorithmic}
\usepackage{hyperref}

\usepackage{url}            % simple URL typesetting
\usepackage{booktabs}       % professional-quality tables
\usepackage{amsfonts}       % blackboard math symbols
\usepackage{nicefrac}       % compact symbols for 1/2, etc.
\usepackage{microtype}      % microtypography
\usepackage{color}
\usepackage{amsmath}
\usepackage{bm}
\usepackage{amsthm}
\usepackage{times}
\usepackage{amssymb}
\usepackage{xr}
\usepackage{lscape}
\usepackage{dsfont}
\usepackage{authblk}

\def\nn{\nonumber}

\def\E{\mathbb{E}}
\def\mb{\mathbb}
\def\defeq{\triangleq}
\def\mc{\mathcal}

\def\LM{\mathrm{LM}}

\allowdisplaybreaks

\title{Unsupervised Sequence Classification using Sequential Output Statistics}

\author[]{\bf Yu Liu\thanks{All the authors contributed equally to the paper.}~}
\author[$*$]{\bf Jianshu Chen~}
\author[$*$]{\bf Li Deng}
\affil[$*$]{ Microsoft Research, Redmond, WA 98052, USA \\
\texttt{\{t-liuyu, jianshuc,deng\}@microsoft.com}}

\begin{document}

\maketitle

\begin{abstract}
We consider learning a sequence classifier without labeled data by using sequential output statistics. The problem is highly valuable since obtaining labels in training data is often costly, while the sequential output statistics (e.g., language models) could be obtained independently of input data and thus with low or no cost. To address the problem, we propose an unsupervised learning cost function and study its properties. We show that, compared to earlier works, it is less inclined to be stuck in trivial solutions and avoids the need for a strong generative model. Although it is harder to optimize in its functional form, a stochastic primal-dual gradient method is developed to effectively solve the problem. Experiment results on real-world datasets demonstrate that the new unsupervised learning method gives drastically lower errors than other baseline methods. Specifically, it reaches test errors about twice of those obtained by fully supervised learning.
\end{abstract}

\section{Introduction}

Unsupervised learning is one of the most challenging problems in machine learning. It is often formulated as the modeling of how the world works without requiring a huge amount of human labeling effort, e.g. \cite{LeCun2016}. To reach this grand goal, it is necessary to first solve a sub-goal of unsupervised learning with high practical value; that is, learning to predict output labels from input data without requiring costly labeled data. Toward this end, we study in this paper the learning of a sequence classifier without labels by using sequential output statistics. The problem is highly valuable since the sequential output statistics, such as language models, could be obtained independently of the input data and thus with no labeling cost.

The problem we consider here is different from most studies on unsupervised learning, which concern automatic discovery of inherent regularities of the input data to learn their representations \cite{Book_DeepLearning2016,Smolensky_RBM_1986, hinton2006reducing,hinton_dbn_2006,Blei_LDA_JMLR03,Bengio_deep_09,vincent2010stacked,kingma2013auto,goodfellow2014generative,goodfellowTutorial2016}. When these methods are applied in prediction tasks, either the learned representations are used as feature vectors \cite{Le_Unsupervise_ICML12} or the learned unsupervised models are used to initialize a supervised learning algorithm \cite{dahl2012context,hinton2006reducing, Bengio_greedy_NIPS06, mikolov2013efficient, Dai_semisupervise_NIPS15}. In both ways, the above unsupervised methods played an auxiliary role in helping supervised learning when it is applied to prediction tasks.

Recently, various solutions have been proposed to address the input-to-output prediction problem without using labeled training data, all without demonstrated successes \cite{Deng2015Interspeech,sutskever2015towards,Chen2016}. Similar to this work, the authors in \cite{Chen2016} proposed an unsupervised cost that also exploits the sequence prior of the output samples to train classifiers. The power of such a strong prior in the form of language models in unsupervised learning was also demonstrated in earlier studies in \cite{Knight_ACL06,berg2013unsupervised}. 
However, these earlier methods did not perform well in practical prediction tasks with real-world data without using additional strong generative models. Possible reasons are inappropriately formulated cost functions and inappropriate choices of optimization methods. For example, it was shown in \cite{Chen2016} that optimizing the highly non-convex unsupervised cost function could easily get stuck in trivial solutions, although adding a special regularization mitigated the problem somewhat.

The solution provided in this paper fundamentally improves these prior works in \cite{Deng2015Interspeech,sutskever2015towards,Chen2016} in following aspects. First, we propose a novel cost function for unsupervised learning, and find that it has a desired coverage-seeking property that makes the learning algorithm less inclined to be stuck in trivial solutions than the cost function in \cite{Chen2016}. Second, we develop a special empirical formulation of this cost function that avoids the need for a strong generative model as in \cite{sutskever2015towards}. Third, although the proposed cost function is more difficult to optimize in its functional form, we develop a stochastic primal-dual gradient (SPDG) algorithm to effectively solve problem. Our analysis of SPDG demonstrates how it is able to reduce the high barriers in the cost function by transforming it into a primal-dual domain. Finally and most importantly, we demonstrate the new cost function and the associated SPDG optimization algorithm work well in two real-world classification tasks. In the rest of the paper, we proceed to demonstrate these points and discuss related works along the way.

% =====================================================================

\section{Empirical-ODM: An unsupervised learning cost for sequence classifiers}
\label{Sec:Cost}

In this section, we extend the earlier work of \cite{sutskever2015towards}  and propose an unsupervised learning cost named Empirical Output Distribution Match (Empirical-ODM) for training classifiers without labeled data. We first formulate the unsupervised learning problem with sequential output structures. Then, we introduce the Empirical-ODM cost and discuss its important properties that are closely related to unsupervised learning.

\subsection{Problem formulation}

We consider the problem of learning a sequence classifier that predicts an output sequence $(y_1,\ldots,y_{T_0})$ from an input sequence $(x_1,\ldots,x_{T_0})$ without using labeled data, where $T_0$ denotes the length of the sequence. Specifically, the learning algorithm does not have access to a labeled training set $\mc{D}_{XY} \defeq \{(x_1^n,\ldots,x_{T_n}^n), (y_1^n,\ldots,y_{T_n}^n): n=1,\ldots,M\}$, where $T_n$ denotes the length of the $n$-th sequence. Instead, what is available is a collection of input sequences, denoted as $\mc{D}_X \defeq \{(x_1^n,\ldots,x_{T_n}^n): n=1,\ldots,M\}$. In addition, we assume that the sequential output statistics (or sequence prior), in the form of an $N$-gram probability, are available:
	\begin{align}
		p_{\LM}(i_1,\ldots,i_N) 
				&\defeq 
						p_{\LM}(y^n_{t-N+1}=i_1,\ldots,y^n_{t}=i_N)
				\nn
	\end{align}
where $i_1,\ldots,i_N \in \{1,\ldots,C\}$ and the subscript ``LM'' stands for language model. Our objective is to train the sequence classifier by just using $\mc{D}_X$ and $p_{\LM}(\cdot)$. Note that the sequence prior $p_{\LM}(\cdot)$, in the form of language models, is a type of structure commonly found in natural language data, which can be learned from a large amount of text data freely available without labeling cost. For example, in optical character recognition (OCR) tasks, $y_t^n$ could be an English character and $x_t^n$ is the input image containing this character. We can estimate an $N$-gram character-level language model $p_{\LM}(\cdot)$ from a separate text corpus. Therefore, our learning algorithm will work in a fully unsupervised manner, without any human labeling cost. In our experiment section, we will demonstrate the effectiveness of our method on such a real OCR task. Other potential applications include speech recognition, machine translation, and image/video captioning. 

In this paper, we focus on the sequence classifier in the form of $p_{\theta}(y_t^n|x_t^n)$ that is, it computes the posterior probability $p_{\theta}(y_t^n|x_t^n)$ only based on the current input sample $x_t^n$ in the sequence. Furthermore, we restrict our choice of $p_{\theta}(y_t^n|x_t^n)$ to be linear classifiers\footnote{$p_{\theta}(y_t^n = i | x_t^n) = {e^{\gamma w_i^T x_t^n }}/{\sum_{j=1}^C e^{\gamma w_j^T x_t^n}}$, where the model parameter is $\theta \defeq \{ w_i \in \mb{R}^d, i=1,\ldots, C\}$.} and focus our attention on designing and understanding unsupervised learning costs and methods for label-free prediction. In fact, as we will show in later sections, even with linear models, the unsupervised learning problem is still highly nontrivial and the cost function is also highly non-convex. And we emphasize that developing a successful unsupervised learning approach for linear classifiers, as we do in this paper, provides important insights and is an important first step towards more advanced nonlinear models (e.g., deep neural networks). We expect that, in future work, the insights obtained here could help us generalize our techniques to nonlinear models.

A recent work that shares the same motivations as our work is \cite{Stewart2017}, which also recognizes the high cost of obtaining labeled data and seeks label-free prediction. Different from our setting, they exploit domain knowledge from laws of physics in computer vision applications, whereas our approach exploits sequential statistics in the natural language outputs. Finally, our problem is fundamentally different from the sequence transduction method in \cite{graves2012sequence}, although it also exploits language models for sequence prediction. Specifically, the method in \cite{graves2012sequence} is a fully supervised learning in that it requires supervision at the sequence level; that is, for each input sequence, a corresponding output sequence (of possibly different length) is provided as a label. The use of language model in \cite{graves2012sequence} only serves the purpose of regularization in the sequence-level \emph{supervised} learning. In stark contrast, the unsupervised learning we propose does not require supervision at any level including specifically the sequence level; we do not need the sequence labels but only the prior distribution $p_{\LM}(\cdot)$ of the output sequences.

\subsection{The Empirical-ODM}
\label{Sec:Cost:EmpiricalODM}

We now introduce an unsupervised learning cost that exploits the sequence structure in $p_{\LM}(\cdot)$. It is mainly inspired by the approach to breaking the Caesar cipher, one of the simplest forms of encryption \cite{luciano1987cryptology}. Caesar cipher is a substitution cipher where each letter in the original message is replaced with a letter corresponding to a certain number of letters up or down in the alphabet. For example, the letter ``D'' is replaced by the letter ``A'', the letter ``E'' is replaced by the letter ``B'', and so on. In this way, the original message that was readable ends up being less understandable. The amount of this shifting is also known to the intended receiver of the message, who can decode the message by shifting back each letter in the encrypted message. However, Caesar cipher could also be broken by an unintended receiver (not knowing the shift) when it analyzes the frequencies of the letters in the encrypted messages and matches them up with the letter distribution of the original text \cite[pp.9-11]{beutelspacher1994cryptology}. More formally, let $y_t=f(x_t)$ denote a function that maps each encrypted letter $x_t$ into an original letter $y_t$. And let $p_{\LM}(i) \defeq p_{\LM}(y_t=i)$ denote the prior letter distribution of the original message, estimated from a regular text corpus. When $f(\cdot)$ is constructed in a way that all mapped letters $\{y_t: y_t = f(x_t), t=1,\ldots,T\}$ have the same distribution as the prior $p_{\LM}(i)$, it is able to break the Caesar cipher and recover the original letters at the mapping outputs.

Inspired by the above approach,  the posterior probability $p_{\theta}(y_t^n|x_t^n)$ in our classification problem can be interpreted as a stochastic mapping, which maps each input vector $x_t^n$ (the ``encrypted letter'') into an output vector $y_t^n$ (the ``original letter'') with probability $p_{\theta}(y_t^n|x_t^n)$. Then in a samplewise manner, each input sequence $(x_1^n, \ldots, x_{T_n}^n)$ is stochastically mapped into an output sequence $(y_1^n, \ldots, y_{T_n}^n)$. We move a step further than the above approach by requiring that the distribution of the $N$-grams among all the mapped output sequences are close to the prior $N$-gram distribution $p_{\LM}(i_1,\ldots,i_N)$. With this motivation, we propose to learn the classifier $p_{\theta}(y_t|x_t)$ by minimizing the negative cross entropy between the prior distribution and the expected $N$-gram frequency of the output sequences:
	\begin{align}
		\min_{\theta}
		&\bigg\{
		\mc{J}(\theta)
		\defeq
		-\sum_{i_1,\ldots,i_N} \!\!\!
		p_{\LM} (i_1,\ldots,i_N)
		\ln \overline{p}_{\theta}(i_1,\ldots,i_N)
		\bigg\}
		\label{Equ:ProblemFormulation:NewCost}
	\end{align}
where $\overline{p}_{\theta}(i_1,\ldots,i_N)$ denotes the expected $N$-gram frequency of all the output sequences. In Appendix \ref{Appendix:ExpectedNgramFrequency} of the supplementary material, we derive its expression as
	\begin{align}
		\overline{p}_{\theta}(i_1,\ldots,i_N)
			&\defeq
					\frac{1}{T} \sum_{n=1}^M \sum_{t=1}^{T_n} \prod_{k=0}^{N-1}
					p_{\theta}(y_{t-k}^n = i_{N-k}  | x_{t-k}^n)
		\label{Equ:ProblemFormulation:p_bar_theta}
	\end{align}
where $T \defeq T_1 + \cdots + T_M$ is the total number of samples in all sequences. Note that minimizing the negative cross entropy in \eqref{Equ:ProblemFormulation:NewCost} is also equivalent to minimizing the Kullback-Leibler (KL) divergence between the two distributions since they only differ by a constant term, $\sum p_{\LM} \ln p_{\LM}$. Therefore, the cost function \eqref{Equ:ProblemFormulation:NewCost} seeks to estimate $\theta$ by matching the two output distributions, where the expected $N$-gram distribution in \eqref{Equ:ProblemFormulation:p_bar_theta} is an empirical average over all the samples in the training set.  For this reason, we name the cost \eqref{Equ:ProblemFormulation:NewCost} as \emph{Empirical Output Distribution Match} (Empirical-ODM) cost. 

In \cite{sutskever2015towards}, the authors proposed to minimize an output distribution match (ODM) cost, defined as the KL-divergence between the prior output distribution and the marginalized output distribution, $D(p_{\LM}(y) || p_{\theta}(y))$, where $p_{\theta}(y) \defeq \int p_{\theta}(y|x) p(x) dx$. However, evaluating $p_{\theta}(y)$ requires integrating over the input space using a generative model $p(x)$. Due to the lack of such a generative model, they were not able to optimize this proposed ODM cost. Instead, alternative approaches such as Dual autoencoders and GANs were proposed as heuristics. Their results were not successful without using a few labeled data. Our proposed Empirical-ODM cost is different from the ODM cost in \cite{sutskever2015towards} in three key aspects. (i) We do not need any labeled data for training. (ii) We exploit sequence structure of output statistics, i.e., in our case $y = (y_1,\ldots,y_N)$ ($N$-gram) whereas in \cite{sutskever2015towards} $y = y_t$ (unigram, i.e., no sequence structure). This is crucial in developing a working unsupervised learning algorithm. The change from unigram to $N$-gram allows us to explicitly exploit the sequence structures at the output, which makes the technique from non-working to working (see Table \ref{Tab:ImpactOfLM} in Section \ref{Sec:Experiments}). It might also explain why the method in \cite{sutskever2015towards} failed as it does not exploit the sequence structure. (iii) We replace the marginalized distribution $p_\theta(y)$ by the expected $N$-gram frequency in \eqref{Equ:ProblemFormulation:p_bar_theta}. This is critical in that it allows us to directly minimize the divergence between two output distributions without the need for a generative model, which  \cite{sutskever2015towards} could not do. In fact, we can further show that $\overline{p}_{\theta}(i_1,\ldots,i_N)$ is an empirical approximation of $p_{\theta}(y)$ with $y=(y_1,\ldots,y_N)$ (see Appendix \ref{Appendix:ExpectedNgramFrequency:PYempirical} of the supplementary material). In this way, our cost \eqref{Equ:ProblemFormulation:NewCost} can be understood as an \emph{$N$-gram} and \emph{empirical} version of the ODM cost except for an additive constant, i.e., $y$ is replaced by $y=(y_1,\ldots,y_N)$ and $p_{\theta}(y)$ is replaced by its empirical approximation.

\subsection{Coverage-seeking versus mode-seeking}

We now discuss an important property of the proposed Empirical-ODM cost \eqref{Equ:ProblemFormulation:NewCost} by comparing it with the cost proposed in \cite{Chen2016}. We show that the Empirical-ODM cost has a \emph{coverage-seeking} property, which makes it more suitable for unsupervised learning than the \emph{mode-seeking} cost in \cite{Chen2016}. 

In \cite{Chen2016}, the authors proposed the expected negative log-likelihood as the unsupervised learning cost function that exploits the output sequential statistics. The intuition was to maximize the aggregated log-likelihood of all the output sequences assumed to be generated by the stochastic mapping $p_{\theta}(y_t^n|x_t^n)$. We show in Appendix \ref{Appendix:EquivFormChen2016} of the supplementary material that their cost is equivalent to
	\begin{align}
		-\sum_{i_1,\ldots, i_{N-1}}  
		\sum_{i_N} \;
		\overline{p}_{\theta}(i_1,\ldots,i_N) \ln p_{\LM}(i_N | i_{N-1}, \ldots, i_1)
		\label{Equ:ProblemFormulation:OldCost_Form2}
	\end{align}
where $p_{\LM}(i_N | i_{N-1}, \ldots, i_1) \defeq p(y_t^n=i_N | y_{t-1}^n=i_{N-1},\ldots, y_{t-N+1}^n=i_1)$, and the summations are over all possible values of $i_1,\ldots,i_N \in \{1,\ldots,C\}$. In contrast, we can rewrite our cost \eqref{Equ:ProblemFormulation:NewCost} as
	\begin{align}		
		-
		&\sum_{i_1,\ldots, i_{N-1}}		
		p_{\LM}(i_1, \ldots, i_{N-1}) 
		\cdot
		\sum_{i_N} p_{\LM}(i_N|i_{N-1},\ldots,i_1)
		\ln \overline{p}_{\theta}(i_1,\ldots,i_N)
		\label{Equ:ProblemFormulation:NewCost_Form0}
	\end{align}
where we used the chain rule of conditional probabilities. Note that both costs \eqref{Equ:ProblemFormulation:OldCost_Form2} and \eqref{Equ:ProblemFormulation:NewCost_Form0} are in a cross entropy form. However, a key difference is that the positions of the distributions $\overline{p}_{\theta}(\cdot)$ and $p_{\LM}(\cdot)$ are swapped.  We show that the cost in the form of \eqref{Equ:ProblemFormulation:OldCost_Form2} proposed in \cite{Chen2016} is a \emph{mode-seeking} divergence between two distributions, while by swapping $\overline{p}_{\theta}(\cdot)$ and $p_{\LM}(\cdot)$, our cost in \eqref{Equ:ProblemFormulation:NewCost_Form0} becomes a \emph{coverage-seeking} divergence (see \cite{minka2005divergence} for a detailed discussion on divergences with these two different behaviors). To understand this, we consider the following two situations:
	\begin{itemize}
		\item
		If $p_{\LM}(i_N | i_{N-1}, \ldots, i_1) \rightarrow 0$ and
		$\overline{p}_{\theta}(i_1,\ldots,i_N)>0$ for a certain $(i_1,\ldots, i_N)$,
		the cross entropy in
		\eqref{Equ:ProblemFormulation:OldCost_Form2} goes to $+\infty$
		and the cross entropy in \eqref{Equ:ProblemFormulation:NewCost_Form0} 
		approaches zero.
		\item
		If $p_{\LM}(i_N | i_{N-1}, \ldots, i_1) >  0$ and
		$\overline{p}_{\theta}(i_1,\ldots,i_N) \rightarrow 0$ for a certain $(i_1,\ldots,i_N)$, 
		the cross entropy in \eqref{Equ:ProblemFormulation:OldCost_Form2} approaches zero
		and the cross entropy in \eqref{Equ:ProblemFormulation:NewCost_Form0} 
		goes to $+\infty$.
	\end{itemize}
Therefore, the cost function \eqref{Equ:ProblemFormulation:OldCost_Form2} will heavily penalize the classifier if it predicts an output that is believed to be less probable by the prior distribution $p_{\LM}(\cdot)$, and it will not penalize the classifier when it does not predict an output that $p_{\LM}(\cdot)$ believes to be probable. That is, the classifier is encouraged to predict a single output mode with high probability in $p_{\LM}(\cdot)$, a behavior called ``mode-seeking'' in \cite{minka2005divergence}. This probably explains the phenomena observed in \cite{Chen2016}: the training process easily converges to a trivial solution of predicting the same output that has the largest probability in $p_{\LM}(\cdot)$. In contrast, the cost \eqref{Equ:ProblemFormulation:NewCost_Form0} will heavily penalize the classifier if it does not predict the output that $p_{\LM}(\cdot)$ is positive, and will penalize less if it predicts outputs that $p_{\LM}(\cdot)$ is zero. That is, this cost will encourage $p_{\theta}(y|x)$ to cover as much of $p_{\LM}(\cdot)$ as possible, a behavior called ``coverage-seeking'' in \cite{minka2005divergence}. Therefore, training the classifier using \eqref{Equ:ProblemFormulation:NewCost_Form0} will make it less inclined to learn trivial solutions than that in \cite{Chen2016} since it will be heavily penalized. We will verify this fact in our experiment section \ref{Sec:Experiments}. In summary, our proposed cost \eqref{Equ:ProblemFormulation:NewCost} is more suitable for unsupervised learning than that in \cite{Chen2016}.

\subsection{The difficulties of optimizing $\mc{J}(\theta)$}
\label{Sec:Cost:Difficulties}

However, there are two main challenges of optimizing the Empirical-ODM cost $\mc{J}(\theta)$ in \eqref{Equ:ProblemFormulation:NewCost}. The first one is that the sample average (over the entire training data set) in the expression of $\overline{p}_{\theta}(\cdot)$ (see \eqref{Equ:ProblemFormulation:p_bar_theta}) is inside the logarithmic loss, which is different from traditional machine learning problems where the average is outside loss functions (e.g., $\sum_t f_t(\theta)$). This functional form prevents us from applying stochastic gradient descent (SGD) to minimize \eqref{Equ:ProblemFormulation:NewCost} as the stochastic gradients would be intrinsically biased (see Appendix \ref{Appendix:AnalysisSGD} for a detailed discussion and see section \ref{Sec:Experiments} for the experiment results). The second challenge is that the cost function $\mc{J}(\theta)$ is highly non-convex even with linear classifiers. To see this, we visualize the profile of the cost function $\mc{J}(\theta)$ (restricted to a two-dimensional sub-space) around the supervised solution in Figure \ref{Fig:Profile_J_Main}.\footnote{The approach to visualizing the profile is explained with more detail in Appendix \ref{Appendix:VisualizationDetails}. More slices and a video of the profiles from many angles can be found in the supplementary material.} We observe that there are local optimal solutions and there are high barriers between the local and global optimal solutions. Therefore, besides the difficulty of having the sample average inside the logarithmic loss,  minimizing this cost function directly will be difficult since crossing the high barriers to reach the global optimal solution would be hard if not properly initialized.

%\subsection{Profiles in the primal-dual domain}
\begin{figure*}[t!]
	\centering
	\begin{subfigure}[t]{0.31\textwidth}
		\includegraphics[width=\textwidth]{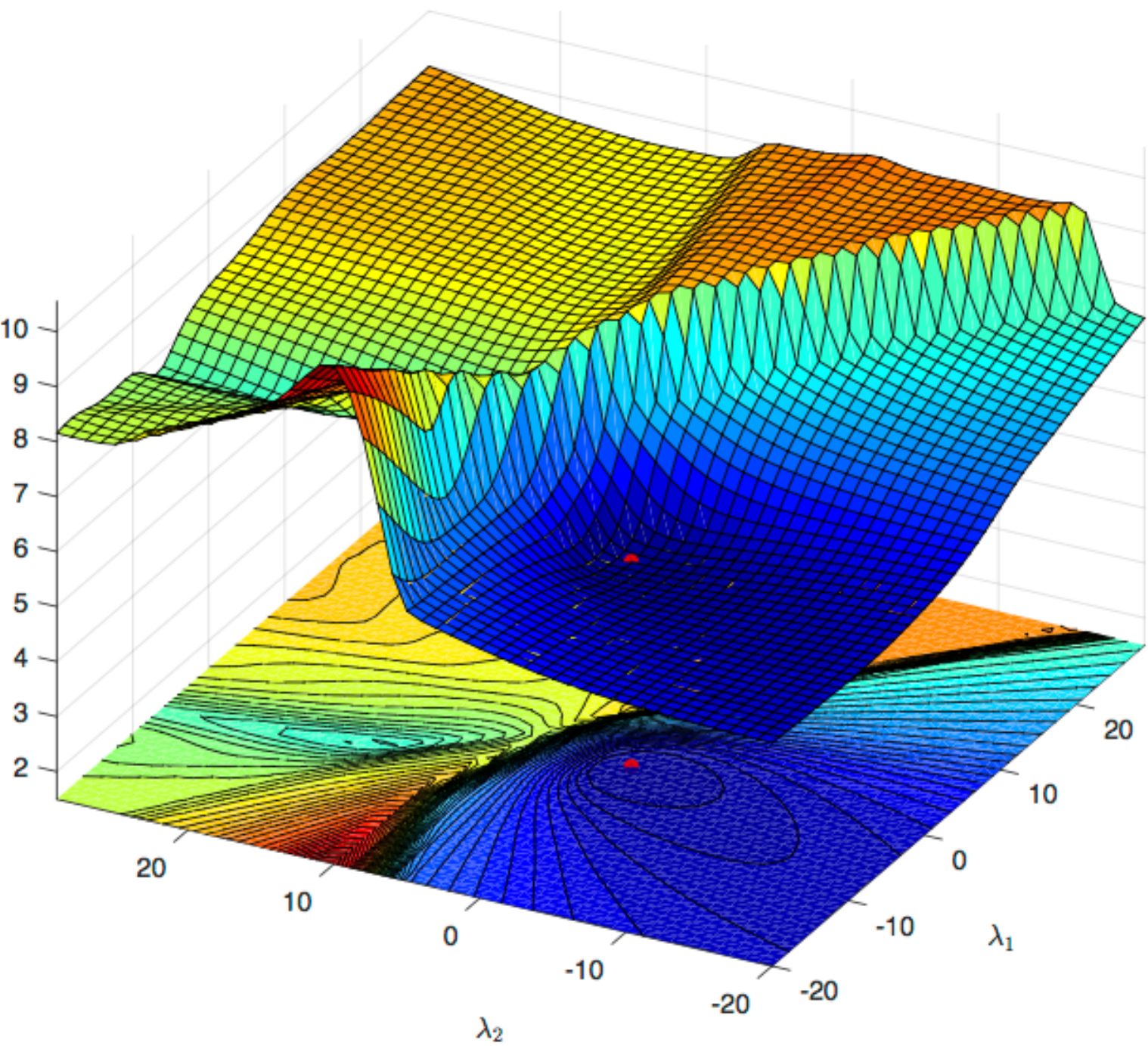}
		\caption{}
	\end{subfigure}	
	\quad
	\begin{subfigure}[t]{0.31\textwidth}
		\includegraphics[width=\textwidth]{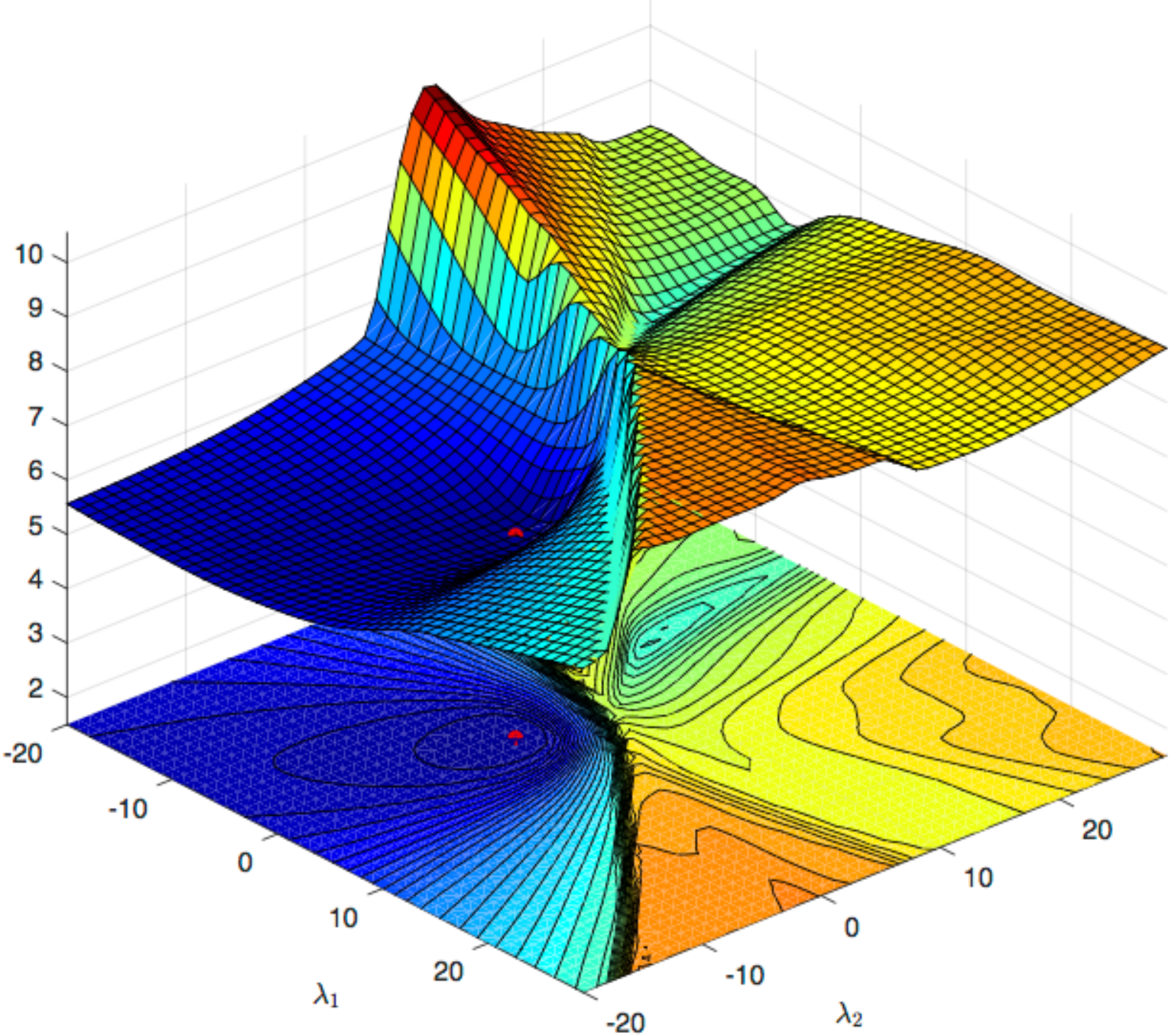}
		\caption{}
	\end{subfigure}
	\quad	
	\begin{subfigure}[t]{0.31\textwidth}
		\includegraphics[width=\textwidth]{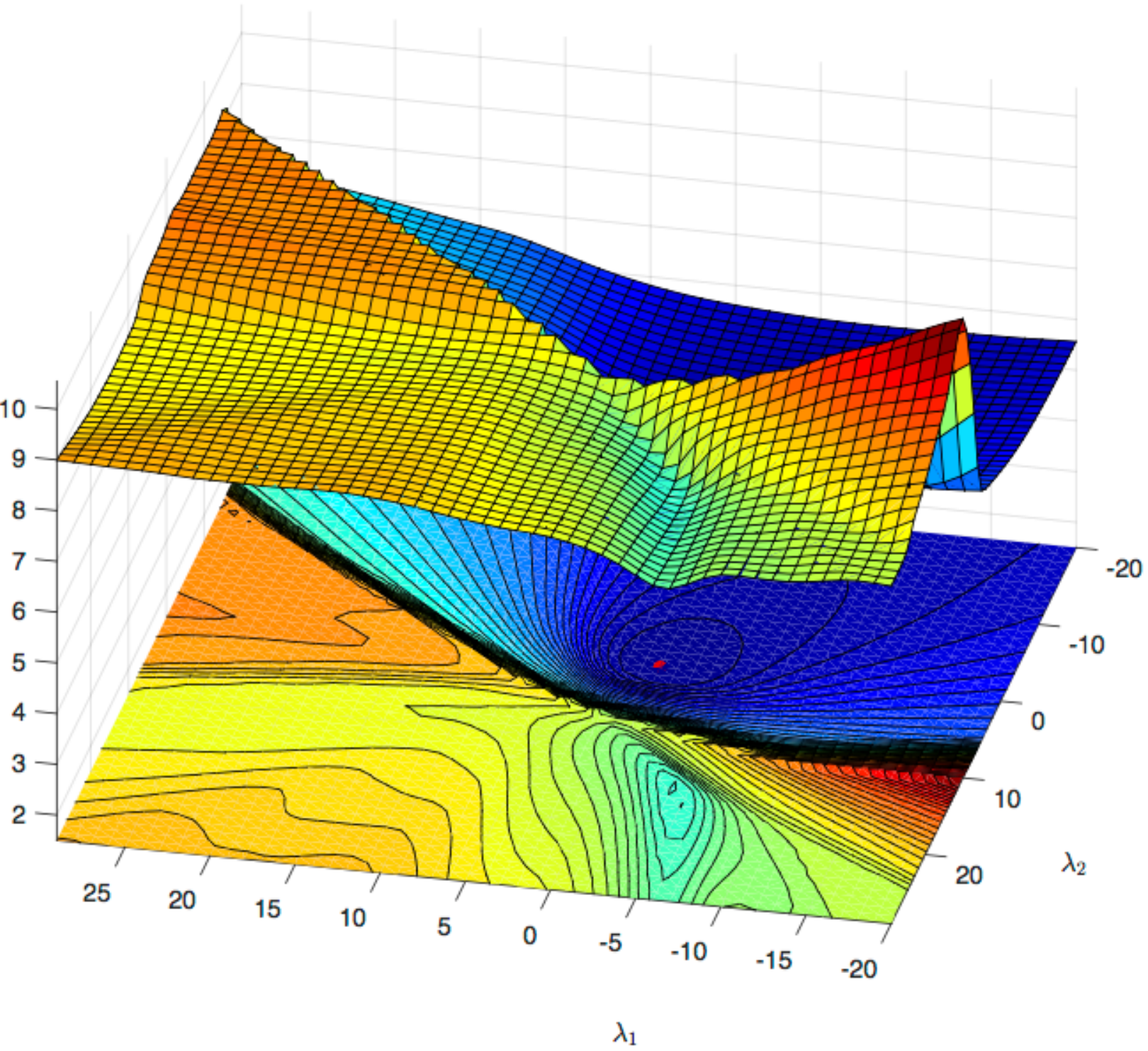}
		\caption{}
	\end{subfigure}	
	\caption{The profiles of $\mc{J}(\theta)$ for the OCR dataset on a two-dimensional affine space passing through the supervised solution. The three figures show the same profile from different angles, where the red dot is the supervised solution. The contours of the profiles are shown at the bottom.
	}
	\label{Fig:Profile_J_Main}
\end{figure*}

\section{The Stochastic Primal-Dual Gradient (SPDG) Algorithm}
\label{Sec:SPDG}

To address the first difficulty in Section \ref{Sec:Cost:Difficulties}, we transform the original cost \eqref{Equ:ProblemFormulation:NewCost} into an equivalent min-max problem in order to bring the sample average out of the logarithmic loss. Then, we could obtain unbiased stochastic gradients to solve the problem. To this end, we first introduce the concept of \emph{convex conjugate functions}. For a given convex function $f(u)$, its convex conjugate function $f^{\star}(\nu)$ is defined as $f^{\star}(\nu) \defeq \sup_u (\nu^T u - f(u))$ \cite[pp.90-95]{boyd2004convex}, where $u$ and $\nu$ are called primal and dual variables, respectively. For a scalar function $f(u) = -\ln u$, its conjugate function can be calculated as $f^{\star}(\nu) = -1 - \ln(-\nu)$ with $\nu<0$. Furthermore, it holds that $f(u) = \sup_{\nu}( u^T \nu - f^{\star}(\nu) )$, by which we have $-\ln u = \max_{\nu} ( u \nu + 1 + \ln(-\nu) )$.\footnote{The supremum is attainable and is thus replaced by maximum.} Substituting it into \eqref{Equ:ProblemFormulation:NewCost}, the original minimization problem becomes the following equivalent min-max problem:
	\begin{align}
		\min_{\theta}&\max_{\{\nu_{i_1,\ldots,i_N<0}\}}
		\bigg\{
			\mc{L}(\theta, V)
			\defeq
					\frac{1}{T} \sum_{n=1}^M\sum_{t=1}^{T_n} L_t^n(\theta, V)
			+ \!\!\!
			\sum_{i_1,\ldots,i_N}  \!\!
			p_{\LM}(i_1,\ldots,i_N) \ln(-\nu_{i_1,\ldots,i_N})
		\bigg\}
		\label{Equ:SPDG:SaddlePointProblem}
	\end{align}
where $V \defeq \{\nu_{i_1,\ldots,i_N}\}$ is a collection of all the dual variables $\nu_{i_1,\ldots,i_N}$, and  $L_t^n(\theta, V)$ is the $t$-th component function in the $n$-th sequence, defined as
	\begin{align}
		L_t^n(\theta,V)
			&\defeq
					\sum_{i_1,\ldots,i_N}	\!\!\!	
					p_{\LM}(i_1,\ldots,i_N)
					\nu_{i_1,\ldots,i_N}
					\prod_{k=0}^{N-1}
					p_{\theta}(y_{t-k}^n \!=\! i_{N-k} | x_{t-k}^n)
					\nn
	\end{align}
In the equivalent min-max problem \eqref{Equ:SPDG:SaddlePointProblem}, we find the optimal solution $(\theta^{\star}, V^{\star})$ by minimizing $\mc{L}$ with respect to the \emph{primal} variable $\theta$ and maximizing $\mc{L}$ with respect to the \emph{dual} variable $V$. The obtained optimal solution to \eqref{Equ:SPDG:SaddlePointProblem}, $(\theta^{\star}, V^{\star})$,  is called the saddle point of $\mc{L}$ \cite{boyd2004convex}. Once it is obtained, we only keep $\theta^{\star}$, which is also the optimal solution to \eqref{Equ:ProblemFormulation:NewCost} and thus the model parameter.

\begin{algorithm}[tb]
	\renewcommand{\algorithmicrequire}{\textbf{Inputs:}}
	\renewcommand{\algorithmicensure}{\textbf{Outputs:}}
	\caption{Stochastic Primal-Dual Gradient Method}
	\label{Alg:SPDG}
	\begin{algorithmic}[1]	
		\STATE {\bf Input data:} $\mc{D}_X = \{(x_1^n,\ldots, x_{T_n}^n): n=1,\ldots,M\}$ and $p_{\LM}(i_1,\ldots,i_N)$.
		\STATE Initialize $\theta$ and $V$ where the elements of $V$ are negative
		\REPEAT
			\STATE Randomly sample a mini-batch of $B$ subsequences of length $N$ from
			all the sequences in the training set $\mc{D}_X$, i.e.,
			$\mc{B} = \{ (x_{t_m-N+1}^{n_m},\ldots, x_{t_m}^{n_m})\}_{m=1}^B$.
			\STATE Compute the stochastic gradients for each subsequence
			in the mini-batch and average them
			{\scriptsize
				\begin{align}
					\Delta \theta
						&=	
							\frac{1}{B} \sum_{m=1}^B \frac{\partial L_{t_m}^{n_m}}{\partial \theta},
							\quad							
					\Delta V
						=
							\frac{1}{B} \sum_{m=1}^B \!\! \frac{\partial L_{t_m}^{n_m}}{\partial V}
							\!+\!
							\frac{\partial}{\partial V} \!\!\!
							\sum_{i_1\!\ldots\!i_N}  \!\!\!\!
							p_{\LM}(i_1,\!\ldots\!,i_N) \ln(\! -\nu_{i_1,\ldots,i_N} \!)
							\nn
				\end{align}
			}
			\STATE Update $\theta$ and $V$ according to 
			$\theta \leftarrow \theta - \mu_{\theta} \Delta \theta$ and
			$V \leftarrow V + \mu_v \Delta V$.
		\UNTIL{convergence or a certain stopping condition is met}
	\end{algorithmic}
\end{algorithm}
	
We further note that the equivalent min-max problem \eqref{Equ:SPDG:SaddlePointProblem} is now in a form that sums over $T=T_1+\cdots+T_M$ component functions $L_t^n(\theta, V)$. Therefore, the empirical average has been brought out of the logarithmic loss and we are ready to apply stochastic gradient methods. Specifically, we  minimize $\mc{L}$ with respect to the \emph{primal} variable $\theta$ by stochastic gradient \emph{descent} and maximize $\mc{L}$ with respect to the \emph{dual} variable $V$ by stochastic gradient \emph{ascent}. Therefore, we name the algorithm stochastic primal-dual gradient (SPDG) method (see its details in Algorithm \ref{Alg:SPDG}). We implement the SPDG algorithm in TensorFlow, which automatically computes the stochastic gradients.\footnote{The code will be released soon.} Finally, the constraint on dual variables $\nu_{i_1,\ldots,i_N}$ are automatically enforced by the inherent log-barrier, $\ln(-\nu_{i_1,\ldots,i_N})$, in \eqref{Equ:SPDG:SaddlePointProblem} \cite{boyd2004convex}. Therefore, we do not need a separate method to enforce the constraint.

\begin{figure*}[t!]
	\centering
	\begin{subfigure}[t]{0.44\textwidth}
		\includegraphics[width=\textwidth]{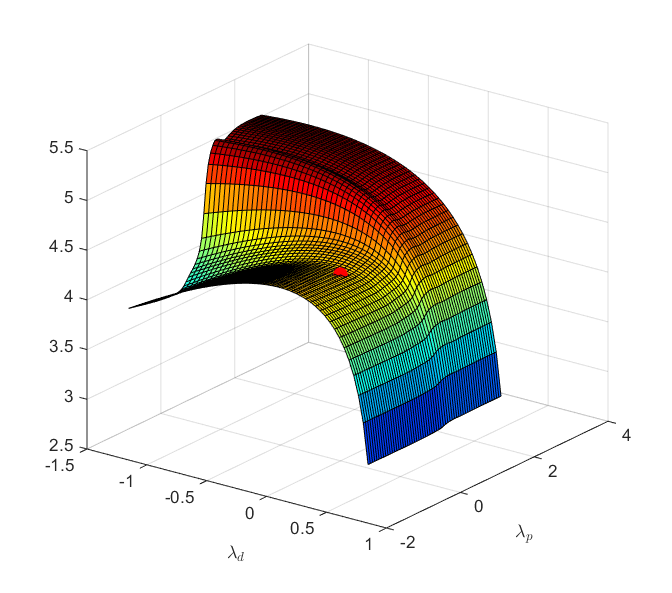}
		\caption{}
		\label{Fig:Profile_L_Main:2D}
	\end{subfigure}	
	\qquad
	\begin{subfigure}[t]{0.46\textwidth}
		\includegraphics[width=\textwidth]{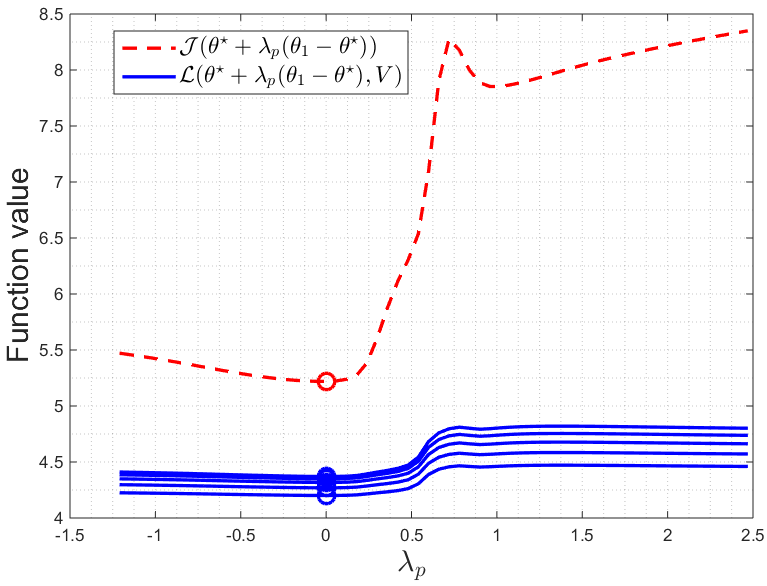}
		\caption{}
		\label{Fig:Profile_L_Main:1D}
	\end{subfigure}
	\caption{The profiles of $\mc{L}(\theta, V)$ for the OCR dataset. (a) The profile on a two-dimensional affine space passing through the optimal solution (red dot). (b) The profile along the line of $\theta^{\star} + \lambda_p (\theta_1-\theta^{\star})$ with $\lambda_p \in \mb{R}$, where the circles are the optimal solutions.
	}
	\label{Fig:Profile_L_Main}
\end{figure*}

We now show that the above min-max (primal-dual) reformulation also alleviates the second difficulty discussed in Section \ref{Sec:Cost:Difficulties}. Similar to the case of $\mc{J}(\theta)$, we examine the profile of $\mc{L}(\theta,V)$ in \eqref{Equ:SPDG:SaddlePointProblem} (restricted to a two-dimensional sub-space) around the optimal (supervised) solution in Figure \ref{Fig:Profile_L_Main:2D} (see Appendix \ref{Appendix:VisualizationDetails} for the visualization details). Comparing Figure \ref{Fig:Profile_L_Main:2D} to Figure \ref{Fig:Profile_J_Main}, we observe that the profile of $\mc{L}(\theta,V)$ is smoother than that of $\mc{J}(\theta)$ and the barrier is significantly lower. To further compare $\mc{J}(\theta)$ and $\mc{L}(\theta,V)$, we plot in Figure \ref{Fig:Profile_L_Main:1D} the values of $\mc{J}(\theta)$ and $\mc{L}(\theta, V)$ along the same line of $\theta^{\star} + \lambda_p (\theta_1-\theta^{\star})$ for different $\lambda_p$. It shows that the barrier of $\mc{L}(\theta, V)$ along the primal direction is lower than that in $\mc{J}(\theta)$. These observations imply that the reformulated min-max problem \eqref{Equ:SPDG:SaddlePointProblem} is better conditioned than the original problem \eqref{Equ:ProblemFormulation:NewCost}, which further justifies the use of SPDG method.

%
%\section{Profiles of the Cost Functions}
%\label{Sec:Profile}

%In this section, we analyze the profiles of the proposed cost functions $\mc{J}(\theta)$ in \eqref{Equ:ProblemFormulation:NewCost} and the function $\mc{L}(\theta,V)$ in \eqref{Equ:SPDG:SaddlePointProblem}. We will show below the difficulty of optimizing the original cost function $\mc{J}(\theta)$ and how the primal-dual reformulation makes the problem much easier.

%\subsection{Profiles in the primal domain}

%=====================================================================

\section{Experiments}
\label{Sec:Experiments}

\subsection{Experimental setup}

We evaluate our unsupervised learning scheme described in earlier secitons using two classification tasks, unsupervised character-level OCR and unsupervised English Spelling Correction (Spell-Corr). In both tasks, there is no label provided during training. Hence, they are both unsupervised.

%In OCR task, images with printed English language symbols are input and we learn to recognize the true characters with out any human labels. 

For the OCR task, we obtain our dataset from a public database UWIII English Document Image Database \cite{UWIII}, which contains images for each line of text with its corresponding groudtruth.  We first use Tesseract \cite{Tesseract} to segment the image for each line of text into characters tiles and assign each tile with one character. We verify the segmentation result by training a simple neural network classifier on the segmented results and achieve 0.9\% error rate on the test set. Then, we select sentence segments that are longer than 100 and contain only lowercase English characters and common punctuations (space, comma, and period). As a result, we have a vocabulary of size 29 and we obtain 1,175 sentence segments including 153,221 characters for our OCR task. To represent images, we extract VGG19 features with $dim=4096$, and project them into $200$-dimension vectors using Principal Component Analysis. We train the language models (LM) $p_{\LM}(\cdot)$ to provide the required output sequence statistics from both in-domain and out-of-domain data sources. The out-of-domain data sources are completely different databases, including three different language partitions (CNA, NYT, XIN) in the English Gigaword database \cite{gigaword}.

In Spell-Corr task, we learn to correct the spelling from a mis-spelled text. From the AFP partition of the Gigaword database, we select 500 sentence segments into our Spell-Corr dataset. We select sentences that are longer than 100 and contain only English characters and common punctuations, resulting in a total of 83,567 characters. The mis-spelled texts are generated by substitution simulations and are treated as our inputs. The objective of this task is to recover the original text.%
% ADD BELOW AFTER THE PAPER IS ACCEPTED (due to anonymity rule)
\footnote{We gratefully acknowledge the discussions with Prof. Hermann Ney for private discussions on this task and his work on using likelihood as the objective function for unsupervised training.}

\subsection{Results: Comparing optimization algorithms}

In the first set of experiments, we aim to evaluate the effectiveness of the SPDG method as described in Section \ref{Sec:SPDG}, which is designed for optimizing the Empirical-ODM cost in Section \ref{Sec:Cost}. The analysis provided in Sections \ref{Sec:Cost} and \ref{Sec:SPDG} sheds insight to why SPDG is superior to the method in \cite{Chen2016} and to the standard stochastic gradient descent (SGD) method. The coverage-seeking behavior of the proposed Empirical-ODM cost helps avoid trivial solutions, and the simultaneous optimization of primal-dual variables reduces the barriers in the highly non-convex profile of the cost function. Furthermore, we do not include the methods from \cite{sutskever2015towards} because their approaches could not achieve satisfactory results without a few labeled data, while we only consider fully unsupervised learning setting. In addition, the methods in \cite{sutskever2015towards} are not optimizing the ODM cost and do not exploit the output sequential statistics.

Table 1 provides strong experimental evidence demonstrating the substantially greater effectiveness of the primal-dual method over the SGD and the method in \cite{Chen2016} on both tasks. All these results are obtained by training the models until converge. Let us examine the results on the OCR in detail. First, the SPGD on the unsupervised cost function achieves 9.21\% error rate, much lower than the error rates of any of mini-batch SGD runs, where the size of the mini-batches ranges from 10 to 10,000.  Note that, larger mini-batch sizes produce lower errors here because it becomes closer to full-batch gradient and thus lower bias in SGD. On the other hand, when the mini-batch size is as small as 10, the high error rate of 83.09\% is close to a guess by majority rule --- predicting the character (space) that has a largest proportion in the train set, i.e., $25,499/153,221 = 83.37\%$. Furthermore,  the method from \cite{Chen2016} does not perform well no matter how we tune the hyperparameters for the generative regularization. Finally and perhaps most interestingly, with no labels provided in the training, the classification errors produced by our method are only about twice compared with supervised learning (4.63\% shown in Table 1). This clearly demonstrates that the unsupervised learning scheme proposed in this paper is an effective one. For the Spelling Correction data set (see the third column in Table \ref{Tab:DifferentAlg}), we observe rather consistent results with the OCR data set.

\begin{table}[ht]
\caption{Test error rates on two datasets: OCR and Spell-Corr. The 2-gram character LM is trained from in-domain data. The numbers inside $\langle \cdot \rangle$ are the mini-batch sizes of the SGD method.}
\label{Tab:DifferentAlg}
\setlength\tabcolsep{2.5pt}
\begin{center}
\begin{tabular}{p{1.8cm}p{1.3cm}p{1.3cm}p{1.2cm}p{1.2cm}p{1.2cm}p{1.2cm}p{1.6cm}p{1.4cm}}
\hline
Data sets  & SPDG (Ours) & Method from \cite{Chen2016} & SGD $\langle 10 \rangle$ & SGD $\langle 100 \rangle$ & SGD $\langle 1k \rangle$  & SGD $\langle 10k \rangle$ & Supervised Learning & Majority Guess \\ 
\hline
OCR &\textbf{9.59\%}& 83.37\% & 83.09\% &78.05\%&67.14\% &56.48\%&4.63\%&83.37\% \\ 
Spell-Corr &{\bf 1.94\%}&82.91\%&82.91\%&72.93\%&65.69\%&45.24\%&0.00\%&82.91\% \\
\hline

% \begin{tabular} {lrrcc}
% \hline
% Data-sets & OCR &  Spell-Corr \\
% \hline
% SPDG (this paper)         & {\bf 9.59\%} &  {\bf 1.94\%} \\
% \hline
% Method from \cite{Chen2016} & 83.37\%	&	82.91\% \\
% \hline
% SGD (miniBatch,10)        & 83.09\%  & 82.91\% \\
% SGD (miniBatch,100)       & 78.05\%  & 72.93\% \\
% SGD (miniBatch,1000) 	  & 67.14\%  & 65.69\% \\
% SGD (miniBatch,10000)     & 56.48\%  & 45.24\% \\
% \hline
% Supervised learning    &    4.63\%   & 0.00\%\\
% Guess by majority rule    &    83.37\% & 82.91\%\\
% \hline
\end{tabular}
\end{center}
\end{table}

\subsection{Results: Comparing orders of language modeling}

In the second set of experiments, we examine to what extent the use of sequential statistics (e.g. 2- and 3-gram LMs) can do better than the uni-gram LM (no sequential information) in unsupervised learning. The unsupervised prediction results are shown in Table \ref{Tab:ImpactOfLM}, using different data sources to estimate N-gram LM parameters. Consistent across all four ways of estimating reliable N-gram LMs, we observe significantly lower error rates when the unsupervised learning exploits 2-gram and 3-gram LM as sequential statistics compared with exploiting the prior with no sequential statistics (i.e. 1-gram). In three of four cases, exploiting a 3-gram LM gives better results than a 2-gram LM. Furthermore, the comparable error rate associated with 3-gram using out-of-domain output character data (10.17\% in Table \ref{Tab:ImpactOfLM}) to that using in-domain output character data (9.59\% in Table \ref{Tab:DifferentAlg}) indicates that the effectiveness of the unsupervised learning paradigm presented in this paper is robust to the quality of the LM acting as the sequential prior.

\begin{table}[ht]
\caption{Test error rates on the OCR dataset. Character-level language models (LMs) with the orders are trained from three out-of-domain datasets and from the fused in-domain and out-of-domain data.}
\label{Tab:ImpactOfLM}
\begin{center}
\begin{small}
\begin{tabular}{ccccc}
\hline
 & \texttt{NYT-LM} & \texttt{XIN-LM} & \texttt{CNA-LM} & \texttt{Fused-LM}\\
\hline
No. Sents    & 1,206,903  & 155,647 & 12,234                   & 15,409\\
No. Chars   & 86,005,542 & 18,626,451 & 1,911,124                & 2,064,345\\
1-gram   		& 71.83\%     & 72.14\% & 71.51\%          & 71.25\%\\
2-gram   		& 10.93\%     & {\bf 12.55}\% & 10.56\%          & 10.33\%\\
3-gram   		& {\bf 10.17}\%     & 12.89\% & {\bf 10.29} \%          & {\bf 9.21}\%\\
\hline
\end{tabular}
\end{small}
\end{center}
\end{table}

\section{Conclusions and future work}

In this paper, we study the problem of learning a sequence classifier without the need for labeled training data. The practical benefit of such unsupervised learning is tremendous. For example, in large scale speech recognition systems, the currently dominant supervised learning methods typically require a few thousand hours of training data, where each utterance in the acoustic form needs to be labeled by humans. Although there are millions of hours of natural speech data available for training, labeling all of them for supervised learning is less feasible. To make effective use of such huge amounts of acoustic data, the practical unsupervised learning approach discussed in this paper would be called for. Other potential applications such as machine translation, image and video captioning could also benefit from our paradigm. This is mainly because of their common natural language output structure, from which we could exploit the sequential structures for learning the classifier without labels. Furthermore, our proposed Empirical-ODM cost function significantly improves over the one in \cite{Chen2016} by emphasizing the coverage-seeking behavior. Although the new cost function has a functional form that is more difficult to optimize, a novel SPDG algorithm is developed to effectively address the problem. An analysis of profiles of the cost functions sheds insight to why SPDG works well and why previous methods failed. Finally, we demonstrate in two datasets that our unsupervised learning method is highly effective, producing only about twice errors as fully supervised learning, which no previous unsupervised learning could produce without additional steps of supervised learning. While the current work is restricted to linear classifiers, we intend to generalize the approach to nonlinear models (e.g., deep neural nets \cite{DNN_SPMagazine12}) in our future work. We also plan to extend our current method from exploiting N-gram LM to exploiting the currently state-of-the-art neural-LM.

\bibliography{ICML2017_UnsupervisedLearning}

\begin{thebibliography}{10}

\bibitem{Bengio_deep_09}
Yoshua Bengio.
\newblock Learning deep architectures for {AI}.
\newblock {\em Foundations and Trends in Machine Learning}, 2(1):1--127,
  January 2009.

\bibitem{Bengio_greedy_NIPS06}
Yoshua Bengio, Pascal Lamblin, Dan Popovici, and Hugo Larochelle.
\newblock Greedy layer-wise training of deep networks.
\newblock In {\em Proceedings of the Advances in Neural Information Processing
  Systems (NIPS)}, pages 153--160, 2007.

\bibitem{berg2013unsupervised}
Taylor Berg-Kirkpatrick, Greg Durrett, and Dan Klein.
\newblock Unsupervised transcription of historical documents.
\newblock In {\em Proceedings of the 51st Annual Meeting of the Association for
  Computational Linguistics}, pages 207--217, 2013.

\bibitem{beutelspacher1994cryptology}
Albrecht Beutelspacher.
\newblock {\em Cryptology}.
\newblock Mathematical Association of America, 1994.

\bibitem{Blei_LDA_JMLR03}
David~M. Blei, Andrew~Y. Ng, and Michael~I. Jordan.
\newblock Latent dirichlet allocation.
\newblock {\em The Journal of Machine Learning Research}, 3:993--1022, March
  2003.

\bibitem{boyd2004convex}
Stephen Boyd and Lieven Vandenberghe.
\newblock {\em Convex optimization}.
\newblock Cambridge university press, 2004.

\bibitem{Chen2016}
Jianshu Chen, Po-Sen Huang, Xiaodong He, Jianfeng Gao, and Li~Deng.
\newblock Unsupervised learning of predictors from unpaired input-output
  samples.
\newblock {\em arXiv:1606.04646}, 2016.

\bibitem{LeCun2016}
Soumith Chintala and Yann LeCun.
\newblock A path to unsupervised learning through adversarial networks.
\newblock In {\em
  https://code.facebook.com/posts/1587249151575490/a-path-to-unsupervised-learning-through-adversarial-networks/},
  2016.

\bibitem{dahl2012context}
George~E Dahl, Dong Yu, Li~Deng, and Alex Acero.
\newblock Context-dependent pre-trained deep neural networks for
  large-vocabulary speech recognition.
\newblock {\em Audio, Speech, and Language Processing, IEEE Transactions on},
  20(1):30--42, 2012.

\bibitem{Dai_semisupervise_NIPS15}
Andrew~M Dai and Quoc~V Le.
\newblock Semi-supervised sequence learning.
\newblock In {\em Proceedings of the Advances in Neural Information Processing
  Systems (NIPS)}, pages 3079--3087, 2015.

\bibitem{Deng2015Interspeech}
Li~Deng.
\newblock Deep learning for speech and language processing.
\newblock In {\em Tutorial at Interspeech Conf, Dresden, Germany,
  https://www.microsoft.com/en-us/research/wp-content/uploads/2016/07/interspeech-tutorial-2015-lideng-sept6a.pdf,
  Aug-Sept}, 2015.

\bibitem{goodfellowTutorial2016}
Ian Goodfellow.
\newblock Generative adversarial nets.
\newblock In {\em Tutorial at NIPS,
  http://www.cs.toronto.edu/~dtarlow/pos14/talks/goodfellow.pdf}, 2016.

\bibitem{Book_DeepLearning2016}
Ian Goodfellow, Yoshua Bengio, and Aaron Courville.
\newblock Deep {L}earning, by {MIT P}ress.
\newblock 2016.

\bibitem{goodfellow2014generative}
Ian Goodfellow, Jean Pouget-Abadie, Mehdi Mirza, Bing Xu, David Warde-Farley,
  Sherjil Ozair, Aaron Courville, and Yoshua Bengio.
\newblock Generative adversarial nets.
\newblock In {\em Proceedings of the Advances in Neural Information Processing
  Systems (NIPS)}, pages 2672--2680, 2014.

\bibitem{graves2012sequence}
Alex Graves.
\newblock Sequence transduction with recurrent neural networks.
\newblock {\em arXiv preprint arXiv:1211.3711}, 2012.

\bibitem{DNN_SPMagazine12}
Geoffrey Hinton, Li~Deng, Dong Yu, George~E Dahl, Abdel-Rahman Mohamed, Navdeep
  Jaitly, Andrew Senior, Vincent Vanhoucke, Patrick Nguyen, Tara~N. Sainath,
  and B.~Kingsbury.
\newblock Deep neural networks for acoustic modeling in speech recognition: The
  shared views of four research groups.
\newblock {\em IEEE Signal Processing Magazine}, 29(6):82--97, November 2012.

\bibitem{hinton_dbn_2006}
Geoffrey~E Hinton, Simon Osindero, and Yee-Whye Teh.
\newblock A fast learning algorithm for deep belief nets.
\newblock {\em Neural computation}, 18(7):1527--1554, 2006.

\bibitem{hinton2006reducing}
Geoffrey~E Hinton and Ruslan~R Salakhutdinov.
\newblock Reducing the dimensionality of data with neural networks.
\newblock {\em Science}, 313(5786):504--507, 2006.

\bibitem{Tesseract}
Anthony Kay.
\newblock Tesseract: An open-source optical character recognition engine.
\newblock {\em Linux Journal}, 2007.

\bibitem{kingma2013auto}
Diederik~P Kingma and Max Welling.
\newblock Auto-encoding variational bayes.
\newblock {\em arXiv preprint arXiv:1312.6114}, 2013.

\bibitem{Knight_ACL06}
Kevin Knight, Anish Nair, Nishit Rathod, and Kenji Yamada.
\newblock Unsupervised analysis for decipherment problems.
\newblock In {\em Proceedings of the COLING/ACL}, pages 499--506, 2006.

\bibitem{Le_Unsupervise_ICML12}
Quoc Le, Marc'Aurelio Ranzato, Rajat Monga, Matthieu Devin, Kai Chen, Greg
  Corrado, Jeff Dean, and Andrew Ng.
\newblock Building high-level features using large scale unsupervised learning.
\newblock In {\em International Conference in Machine Learning}, 2012.

\bibitem{luciano1987cryptology}
Dennis Luciano and Gordon Prichett.
\newblock Cryptology: From caesar ciphers to public-key cryptosystems.
\newblock {\em The College Mathematics Journal}, 18(1):2--17, 1987.

\bibitem{mikolov2013efficient}
Tomas Mikolov, Kai Chen, Greg Corrado, and Jeffrey Dean.
\newblock Efficient estimation of word representations in vector space.
\newblock {\em arXiv preprint arXiv:1301.3781}, 2013.

\bibitem{minka2005divergence}
Tom Minka.
\newblock Divergence measures and message passing.
\newblock Technical report, Technical report, Microsoft Research, 2005.

\bibitem{gigaword}
Robert et~al Parker.
\newblock English gigaword fourth edition ldc2009t13.
\newblock {\em Philadelphia: Linguistic Data Consortium}, 2009.

\bibitem{UWIII}
Ihsin Phillips, Bhabatosh Chanda, and Robert Haralick.
\newblock http://isis-data.science.uva.nl/events/dlia//datasets/uwash3.html.

\bibitem{Smolensky_RBM_1986}
P.~Smolensky.
\newblock Parallel distributed processing: Explorations in the microstructure
  of cognition, vol. 1.
\newblock chapter Information Processing in Dynamical Systems: Foundations of
  Harmony Theory, pages 194--281. 1986.

\bibitem{Stewart2017}
Russell Stewart and Stefano Ermon.
\newblock Label-free supervision of neural networks with physics and domain
  knowledge.
\newblock In {\em Proceedings of AAAI}, 2017.

\bibitem{sutskever2015towards}
Ilya Sutskever, Rafal Jozefowicz, Karol Gregor, Danilo Rezende, Tim Lillicrap,
  and Oriol Vinyals.
\newblock Towards principled unsupervised learning.
\newblock {\em arXiv preprint arXiv:1511.06440}, 2015.

\bibitem{vincent2010stacked}
Pascal Vincent, Hugo Larochelle, Isabelle Lajoie, Yoshua Bengio, and
  Pierre-Antoine Manzagol.
\newblock Stacked denoising autoencoders: Learning useful representations in a
  deep network with a local denoising criterion.
\newblock {\em The Journal of Machine Learning Research}, 11:3371--3408, 2010.

\end{thebibliography}
\bibliographystyle{plain}

\clearpage
\newpage

\section*{\Large Supplementary Material for ``Unsupervised Sequence Classification using Sequential Output Statistics''}

\appendix
\section{Derivation of the equivalent form of the cost in \cite{Chen2016}}
\label{Appendix:EquivFormChen2016}

The cost function in \cite{Chen2016} can be expressed as:
	\begin{align}
		\E\big[- \sum_{n=1}^M \ln p_{\LM}(y_1^n,\ldots,y_{T_n}^n) | x_1^n, \ldots, x_{T_n}^n \big] 
        \label{Equ:CostFrom7}
      \end{align}
We now show how to derive \eqref{Equ:ProblemFormulation:OldCost_Form2} from the above expression. In $N$-gram case, the language model can be written as
	\begin{align}
		p_{\LM}(y_1^n,\ldots,y_{T_n}^n)
			&=
					\prod_{t=1}^{T_n} p_{\LM}(y_t^n | y_{t-1}^n,\ldots, y_{t-N+1}^n)
					\nn
	\end{align}
Substituting the above expression into the cost \eqref{Equ:CostFrom7}, we obtain
	\begin{align}
		\E\big[& - \sum_{n=1}^M \ln p_{\LM}(y_1^n,\ldots,y_{T_n}^n) | x_1^n, \ldots, x_{T_n}^n \big]
					\nn\\
			&=				
					-\sum_{n=1}^M \sum_{(y_1^n,\ldots,y_{T_n}^n)}
					\prod_{t=1}^{T_n} p_{\theta}(y_t^n | x_t^n)
					\ln p_{\LM}(y_1^n,\ldots,y_{T_n}^n)
					\nn\\
			&=										
					-\sum_{n=1}^M \sum_{(y_1^n,\ldots,y_{T_n}^n)}
					p_{\theta}(y_1^n | x_1^n) \cdots p_{\theta}(y_{T_n}^n | x_{T_n}^n)
					 \times
					\sum_{t=1}^{T_n}
					\ln p_{\LM} (y_t^n | y_{t-1}^n, \ldots, y_{t-N+1}^n)
					\nn\\
			&=
					- \sum_{n=1}^M \sum_{t=1}^{T_n} \sum_{(y_1^n,\ldots,y_{T_n}^n)}
					p_{\theta}(y_1^n | x_1^n) \cdots p_{\theta}(y_{T_n}^n | x_{T_n}^n)
% 					\nn\\
% 					\quad 
					\times   \ln p_{\LM} (y_t^n | y_{t-1}^n, \ldots, y_{t-N+1}^n)
					\nn\\
			&=
					- \sum_{n=1}^M \sum_{t=1}^{T_n} \sum_{(y_t^n,\ldots,y_{t-N+1}^n)} \!\!\!\!\!\!\!\!
					p_{\theta}(y_t^n | x_t^n) \cdots p_{\theta}(y_{t-N+1}^n|x_{t-N+1}^n)
% 					\nn\\
% 					\quad 
					\times
					\ln p_{\LM} (y_t^n | y_{t-1}^n,\ldots,y_{t-N+1}^n)
					\nn\\
					&\quad ~~~~~~~~~~~~~~~~~~~~~~~~
                    \times
					\sum_{y_1^n,\ldots,y_{t-N}^n}
					p_{\theta}(y_1^n|x_1^n)
					\cdots
					p_{\theta}(y_{t-N}^n | x_{t-N}^n)
% 					\nn\\
% 					\quad 
                    \times
					\sum_{y_{t+1}^n,\ldots,y_{T_n}^n}
					p_{\theta}(y_{t+1}^n | x_{t+1}^n)
					\cdots
					p_{\theta}(y_{T_n}^n | x_{T_n}^n)
					\nn\\
			&=
					- \sum_{n=1}^M \sum_{t=1}^{T_n} \sum_{(y_t^n,\ldots,y_{t-N+1}^n)} \!\!\!\!\!\!\!\!
					p_{\theta}(y_t^n | x_t^n) \cdots p_{\theta}(y_{t-N+1}^n|x_{t-N+1}^n)
% 					\nn\\
% 					\quad 
					\times
					\ln p_{\LM} (y_t^n | y_{t-1}^n,\ldots,y_{t-N+1}^n)
					\nn\\
			&=
					- \sum_{n=1}^M \sum_{t=1}^{T_n} \sum_{i_1,\ldots, i_N} \!\!\!\!
					p_{\theta}(y_t^n = i_N | x_t) \cdots p_{\theta}(y_{t-N+1}^n=i_1|x_{t-N+1}^n)
% 					\nn\\
% 					\quad 
					\times
					\ln p_{\LM} (y_t^n = i_N | y_{t-1}^n = i_{N-1},\ldots,y_{t-N+1}^n=i_1)
					\nn\\
			&=
					- \sum_{n=1}^M \sum_{t=1}^{T_n} \sum_{i_1,\ldots, i_N} \!\!\!\!
					p_{\theta}(y_t^n = i_N | x_t^n) \cdots p_{\theta}(y_{t-N+1}^n=i_1|x_{t-N+1}^n)
% 					\nn\\
% 					\quad 
					\times
					\ln p_{\LM} (i_N | i_{N-1},\ldots,i_1)
					\nn\\
			&=
					-  \sum_{i_1,\ldots, i_N}  
					\ln p_{\LM} (i_N | i_{N-1},\ldots,i_1)
% 					\nn\\
% 					\quad 
                    \times
					\sum_{n=1}^M \sum_{t=1}^{T_n}
					p_{\theta}(y_t^n = i_N | x_t) \cdots p_{\theta}(y_{t-N+1}^n=i_1|x_{t-N+1}^n)
					\nn\\
			&=
					-  T \sum_{i_1,\ldots, i_N}  
					\ln p_{\LM} (i_N | i_{N-1},\ldots,i_1)
% 					\nn\\
% 					\quad 
                    \times
					\frac{1}{T}
					\sum_{n=1}^M \sum_{t=1}^{T_n}
					p_{\theta}(y_t^n = i_N | x_t^n) \cdots p_{\theta}(y_{t-N+1}^n=i_1|x_{t-N+1}^n)	
					\nn		
	\end{align}

\section{Properties of $\overline{p}_{\theta}(i_1,\ldots,i_N)$}
\label{Appendix:ExpectedNgramFrequency}

\subsection{$\overline{p}_{\theta}(i_1,\ldots,i_N)$ is the expected $N$-gram frequency of all the output sequences}
In this section, we formally derive the following relation, which interprets $\overline{p}_{\theta}(i_1,\ldots,i_N)$ as the expected frequency of $(i_1,\ldots,i_N)$ in the output sequence:
	\begin{align}
		\E_{\prod_{n=1}^M \prod_{t=1}^{T_n} p_{\theta}(y_t^n|x_t^n)}
					\left[
						\frac{n(i_1,\ldots,i_N)}{T}
					\right]
			&=
					\overline{p}_{\theta}(i_N,\ldots,i_1)
					\nn
	\end{align}
where $T \defeq T_1 + \cdots T_M$. Let $(x_1^n,\ldots,x_{T_n}^n)$ be a given $n$-th input training sequence, and let $(y_1^n,\ldots,y_{T_n}^n)$ be a sequence generated according to the posterior $\prod_{t=1}^{T_n} p_{\theta}(y_t^n|x_t^n)$ (which is the classifier). Furthermore, let $\mb{I}_t^n(i_1,\ldots,i_N)$ denote the indicator function of the event $\{y_{t-N+1}^n=i_1,\ldots,y_{t}^n=i_N\}$, and let $n(i_1,\ldots,i_N)$ denote the number of the $N$-gram $(i_1,\ldots,i_N)$ appearing in all the output sequences $\{(y_1^n,\ldots,y_{T_n}^n): n=1,\ldots,M\}$. Then, we have the following relation:
	\begin{align}
		n(i_1,\ldots,i_N)
			&=
					\sum_{n=1}^M \sum_{t=1}^{T_n} \mb{I}_t^n(i_1,\ldots,i_N)
					\nn
	\end{align}
Obviously, $n(i_1,\ldots,i_N)$ is a function of $\{(y_1^n,\ldots,y_{T_n}^n): n=1,\ldots,M\}$ and is thus a random variable. Taking the conditional expectation of the above expression with respect to $\prod_{n=1}^M\prod_{t=1}^{T_n} p_{\theta}(y_t^n|x_t^n)$, we obtain
	\begin{align}
		&\E_{\prod_{n=1}^M\prod_{t=1}^{T_n} p_{\theta}(y_t^n|x_t^n)} [n(i_1,\ldots,i_N)]
					\nn\\					
			&\quad=
					\sum_{n=1}^M \sum_{t=1}^{T_n} 
					\E_{\prod_{n=1}^M\prod_{t=1}^{T_n} p_{\theta}(y_t^n|x_t^n)}[ \mb{I}_t^n(i_1,\ldots,i_N) ]
					\nn\\
			&\quad=
					\sum_{n=1}^M \sum_{t=1}^{T_n} 
					\E_{\prod_{t=1}^{T_n} p_{\theta}(y_t^n|x_t^n)}[ \mb{I}_t^n(i_1,\ldots,i_N) ]
					\nn\\
			&\quad\overset{(a)}{=}
					\sum_{n=1}^M\sum_{t=1}^{T_n} 
					\prod_{k=0}^{N-1}
					p_{\theta}(y_{t-k}^n = i_{N-k}  | x_{t-k}^n)
					\nn
	\end{align}
where step (a) uses the fact that the expectation of an indicator function of an event equals the probability of the event.
Divide both sides by $T$, the right hand side of the above expression becomes $\overline{p}_{\theta}(i_1,\ldots,i_N)$, and we conclude our proof.	
	
\subsection{$\overline{p}_{\theta}(i_N,\ldots,i_1)$ is an empirical approximation of the marginal output $N$-gram probability}
\label{Appendix:ExpectedNgramFrequency:PYempirical}

First, define the marginal $N$-gram probability $p_{\theta}(i_1,\ldots, i_N)$ as
	\begin{align}
		p_{\theta}(i_1,\ldots, i_N)
			&\defeq
					p_{\theta}(y_1=i_1,\ldots,y_N=i_N)
	\end{align}
For simplicity, we consider the case where the input random variables are discrete, taking finite value from a set $\mc{X}$, then $p_{\theta}(i_1,\ldots,i_N)$ can be written as
	\begin{align}	
		p_{\theta}(i_1,\ldots,i_N)
			&=
					\sum_{(x_1,\ldots,x_N) \in \mc{X}^N}
					\prod_{k=1}^{N}
					p_{\theta}(y_{k} = i_{k}  | x_k)
					p(x_1,\ldots,x_N)
	\end{align}
To show that $\overline{p}_{\theta}(i_N,\ldots,i_1)$ is an empirical approximation of $p_{\theta}(i_1,\ldots,i_N)$, it suffices to show that
	\begin{align}	
		\overline{p}_{\theta}(i_1,\ldots,i_N)
			&=
					\sum_{(x_1,\ldots,x_N) \in \mc{X}^N}
					\prod_{k=1}^{N}
					p_{\theta}(y_{k} = i_{k}  | x_k)
					\hat{p}(x_1,\ldots,x_N)
	\end{align}
where $\hat{p}(x_1,\ldots,x_N)$ is the empirical frequency of the $N$-tuple $(x_1,\ldots,x_N)$ in the dataset $\{(x_1^n,\ldots,x_{T_n}^n): n=1,\ldots,M\}$. The result follows in a straightforward manner from the definition of  $\overline{p}_{\theta}(i_1,\ldots,i_N)$:
	\begin{align}
		\overline{p}_{\theta}(i_1,\ldots,i_N)
			&=
					\frac{1}{T}
					\sum_{n=1}^M\sum_{t=1}^{T_n} 
					\prod_{k=0}^{N-1}
					p_{\theta}(y_{t-k}^n = i_{N-k}  | x_{t-k}^n)
					\nn\\
			&=
					\frac{1}{T}
					\sum_{(x_1,\ldots,x_N) \in \mc{X}^N}
					\prod_{k=0}^{N-1}
					p_{\theta}(y_{t-k}^n = i_{N-k}  | x_{t-k}^n=x_{N-k})
					\times
					n(x_1,\ldots,x_N)
					\nn\\
			&=					
					\sum_{(x_1,\ldots,x_N) \in \mc{X}^N}
					\prod_{k=0}^{N-1}
					p_{\theta}(y_{t-k}^n = i_{N-k}  | x_{t-k}^n=x_{N-k})
					\times
					\frac{n(x_1,\ldots,x_N)}{T}
	\end{align}
where $n(x_1,\ldots,x_N)$ denotes the number of $N$-tuple $(x_1,\ldots,x_N)$ in the dataset $\{(x_1^n,\ldots,x_{T_n}^n): n=1,\ldots,M\}$. The second equality is simply re-organizing the summation in the first expression according to the value of $(x_1,\ldots,x_N)$, i.e., accumulating all the terms inside the double-summation with the same value of $(x_1,\ldots,x_N)$ together. Further note that $p_{\theta}(y_{t-k}^n = i_{N-k}  | x_{t-k}^n=x_{N-k})$ is independent of $t$ and $n$ for any given values of $i_{N-k}$ and $x_{N-k}$, so that
	\begin{align}
		\prod_{k=0}^{N-1}
		p_{\theta}(y_{t-k}^n = i_{N-k}  | x_{t-k}^n=x_{N-k})
			&=
					\prod_{k=1}^{N}
					p_{\theta}(y_{k} = i_{k}  | x_k)
	\end{align}
Then, we can conclude the proof by recognizing that $\hat{p}(x_1,\ldots,x_N) = n(x_1,\ldots,x_N)/T$.

\section{Optimizing Empirical-ODM by SGD is intrinsically biased}
\label{Appendix:AnalysisSGD}

In this section, we show that the stochastic gradient of Empirical-ODM is intrinsically biased. To see this, we can express the (full batch) gradient of $\mc{J}(\theta)$ as
	\begin{align}
		\nabla_{\theta} &\mc{J}(\theta)
			=
				-\sum_{i_1,\ldots,i_N}
				p_{\LM}(i_1,\ldots,i_N)
				\frac{
					\frac{1}{T} \sum_{n=1}^M \sum_{t}^{T_n}
					\nabla_{\theta}
					\Big(
					\prod_{k=0}^{N-1}
					p_{\theta}(y_{t-k}^n = i_{N-k} | x_{t-k}^n)
					\Big)
					}
				{					
					\frac{1}{T} \sum_{n=1}^M \sum_{t}^{T_n} \prod_{k=0}^{N-1}
					p_{\theta}(y_{t-k}^n = i_{N-k} | x_{t-k}^n)
				}
		\label{Equ:ProblemFormulation:Grad_J}
	\end{align}
Note that the gradient expression has sample averages in both the numerator and denominator. Therefore, full batch gradient method is less scalable as it needs to go over the entire training set to compute $\nabla_{\theta} \mc{J}(\theta)$ at each update. To apply SGD, we may obtain an unbiased estimate of it by sampling the numerator with a single component while keeping the denominator the same:
	\begin{align}
		- \!\!\!\!\!\!\sum_{i_1,\ldots,i_N} \!\!\!\!\!
				p_{\LM}(i_1,\ldots,i_N)
				\frac{
					\nabla_{\theta}
					\Big(\!
					\prod_{k=0}^{N-1}
					p_{\theta}(y_{t-k}^n \!=\! i_{N-k} | x_{t-k}^n)
					\!
					\Big)
					}
				{	
					\frac{1}{T}
					\sum_{n=1}^M \sum_{t}^{T_n} \!
					\prod_{k=0}^{N-1} \!
					p_{\theta}(y_{t-k}^n \!=\! i_{N-k} | x_{t-k}^n)
				}
				\nn
	\end{align}
However, this implementation is still not scalable as it needs to average over the entire training set at each update to compute the denominator. On the other hand, if we sample both the numerator and the denominator, i.e.,
	\begin{align}
		- \!\!\!\!\sum_{i_1,\ldots,i_N} \!\!\!\!
				p_{\LM}(i_1,\ldots,i_N)
				\frac{
					\nabla_{\theta}
					\Big(\!
					\prod_{k=0}^{N-1}
					p_{\theta}(y_{t-k}^n \!=\! i_{N-k} | x_{t-k}^n)
					\!
					\Big)
					}
				{	
					\prod_{k=0}^{N-1}
					p_{\theta}(y_{t-k}^n \!=\! i_{N-k} | x_{t-k}^n)
				}
				\nn
	\end{align}
then it will be a biased estimate of the gradient \eqref{Equ:ProblemFormulation:Grad_J}. Our experiments in Section \ref{Sec:Experiments} showed that this biased SGD does not perform well on the unsupervised learning problem. 

\section{Experiment Details}

In the experiment, we implement the model with Python 2.7 and Tensorflow 0.12. 

In training of models both on OCR and Spell-Corr task, we initialize the linear model's parameters (primal variable) with $w_{init}=1/dim(x)$ and $\gamma=10$, where $dim(x)$ is the dimension of input. And we initialize the dual parameters $V_{init}$ with uniformly distributed random variables $v \sim U(-1,0)$. We set the learning rate for primal parameter $\mu_\theta=10^{-6}$ and learning rate for dual parameter $\mu_v=10^{-4}$. We use Adam optimization to train our model.

The test set of OCR is generated also from UWIII database,but avoiding overlap with training set. The size of test set of OCR is 15000. Furthermore, the size of the test set of Spell-Corr is also 15000 without overlapping with the training set.

\section{The details of visualizing the high-dimensional cost functions}
\label{Appendix:VisualizationDetails}

Since $\mc{J}(\theta)$ is a high-dimensional function, it is hard to visualize its full profile. Instead, we use the following procedure to partially visualize $\mc{J}(\theta)$. First, since the supervised learning of linear classifiers is a convex optimization problem, from which we could obtain its global optimal solution $\theta^{\star}.$\footnote{Note that, we solve the supervised learning only for the purpose of understanding our proposed unsupervised learning cost $\mc{J}(\theta)$. In our implementation of the unsupervised learning algorithm, we do not use any of the training label information nor supervised learning algorithms.} Then, we randomly generate two parameter vectors $\theta_1$ and $\theta_2$ and plot the two-dimensional function $\mc{J}\big(\theta^{\star} + \lambda_1(\theta_1-\theta^{\star}) + \lambda_2 (\theta_2-\theta^{\star})\big)$ with respect to $\lambda_1,\lambda_2 \in \mb{R}$, which is a slice of the cost function on a two-dimensional plane.

For the profile of $\mc{L}(\theta,V)$ in \eqref{Equ:SPDG:SaddlePointProblem}, similar to the case of $\mc{J}(\theta)$, in order to visualize $\mc{L}(\theta,V)$, we first solve the supervised learning problem to get $\theta^{\star}$. Then we substitute $\theta^{\star}$ into \eqref{Equ:SPDG:SaddlePointProblem} and maximize $\mc{L}(\theta^{\star}, V)$ over $V$ to obtain $V^{\star}=\{\nu_{i_1,\ldots,i_N}^{\star}\}$, where
	$
		\nu_{i_1,\ldots,i_N}^{\star}
			=
					-1\big/{\frac{1}{T} \sum_{n=1}^M \sum_{t=1}^{T_{n}} \prod_{k=0}^{N-1}
					p_{\theta^{\star}}(y_{t-k}^n = i_{N-k} | x_{t-k}^n)}
	$.
We also randomly generate a $(\theta_1,V_1)$ (with the elements of $V_1$ being negative) and plot in Figure \ref{Fig:Profile_L_Main:2D} the values of $\mc{L}(\theta^{\star} + \lambda_p (\theta_1 - \theta^{\star}), V^{\star} + \lambda_d (V_1 - V^{\star}))$ for different $\lambda_p, \lambda_d \in \mb{R}$. Clearly, the optimal solution (red dot) is at the saddle point of the profile.

\section{Additional visualization of $\mc{J}(\theta)$}

\begin{figure*}[t!]
	\centering
	\begin{subfigure}[t]{0.28\textwidth}
		\includegraphics[width=\textwidth]{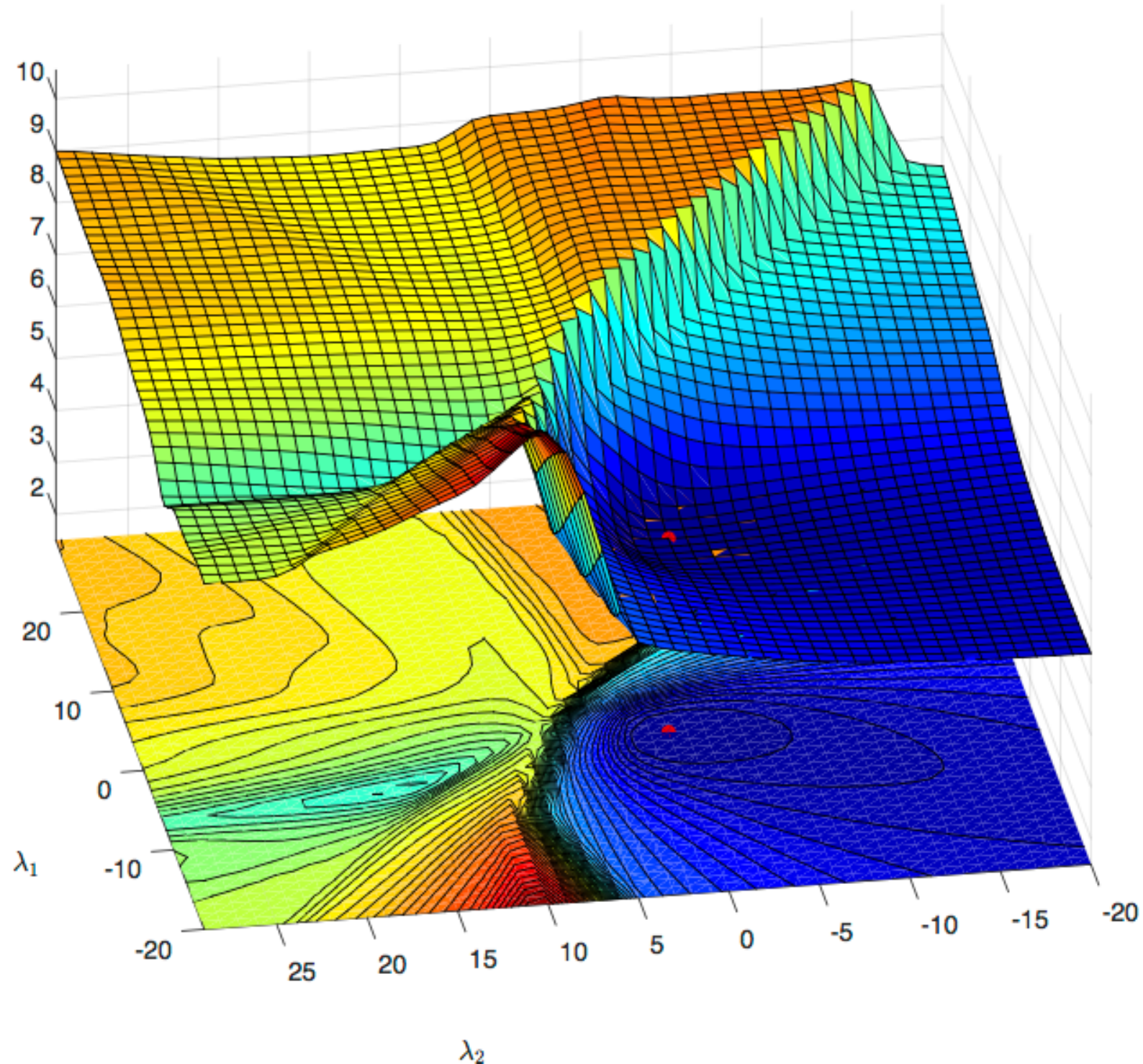}
		\caption{}
	\end{subfigure}	
	\quad
	\begin{subfigure}[t]{0.28\textwidth}
		\includegraphics[width=\textwidth]{figures/surface_OCR_1}
		\caption{}
	\end{subfigure}
	\quad	
	\begin{subfigure}[t]{0.28\textwidth}
		\includegraphics[width=\textwidth]{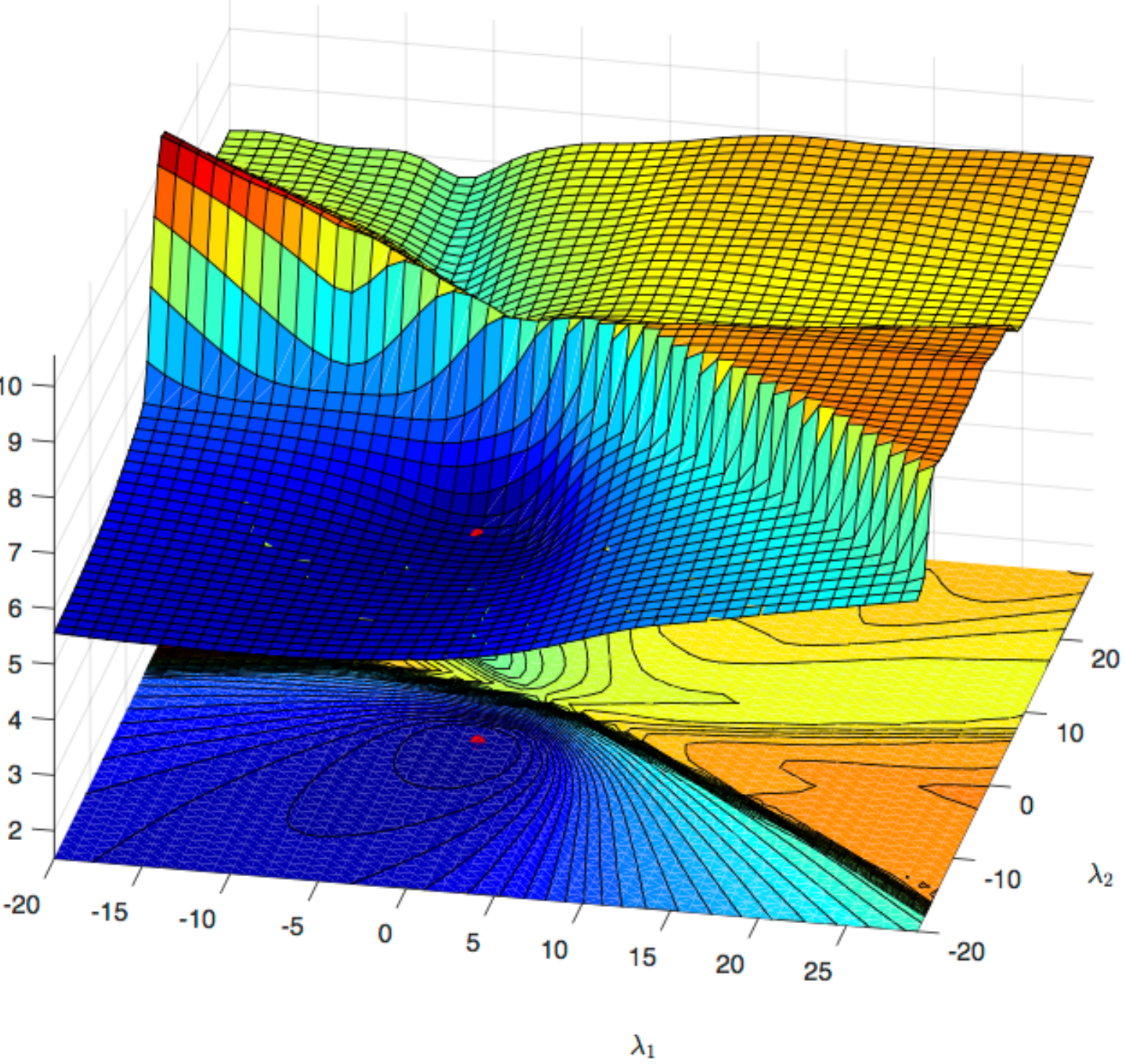}
		\caption{}
	\end{subfigure}	
	\quad
	\begin{subfigure}[t]{0.28\textwidth}
		\includegraphics[width=\textwidth]{figures/surface_OCR_3}
		\caption{}
	\end{subfigure}	
	\quad
	\begin{subfigure}[t]{0.28\textwidth}
		\includegraphics[width=\textwidth]{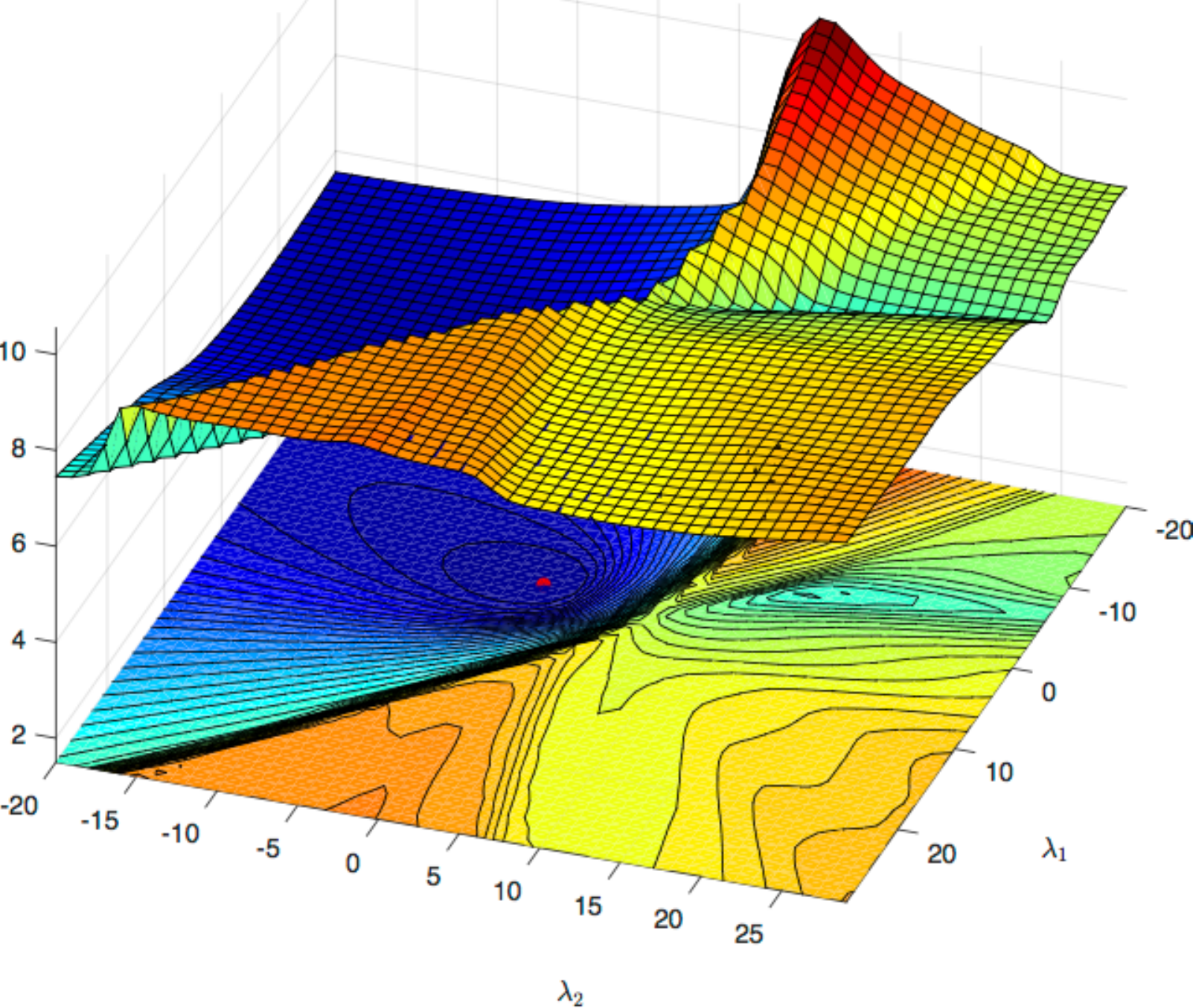}
		\caption{}
	\end{subfigure}
	\quad	
	\begin{subfigure}[t]{0.28\textwidth}
		\includegraphics[width=\textwidth]{figures/surface_OCR_5}
		\caption{}
	\end{subfigure}	
	\caption{Profile Example I: $\mc{J}(\theta)$ for the OCR dataset on a two-dimensional affine space
	}
	\label{Fig:SurfaceOCR1}
\end{figure*}

\begin{figure*}[t!]
	\centering
	\begin{subfigure}[t]{0.28\textwidth}
		\includegraphics[width=\textwidth]{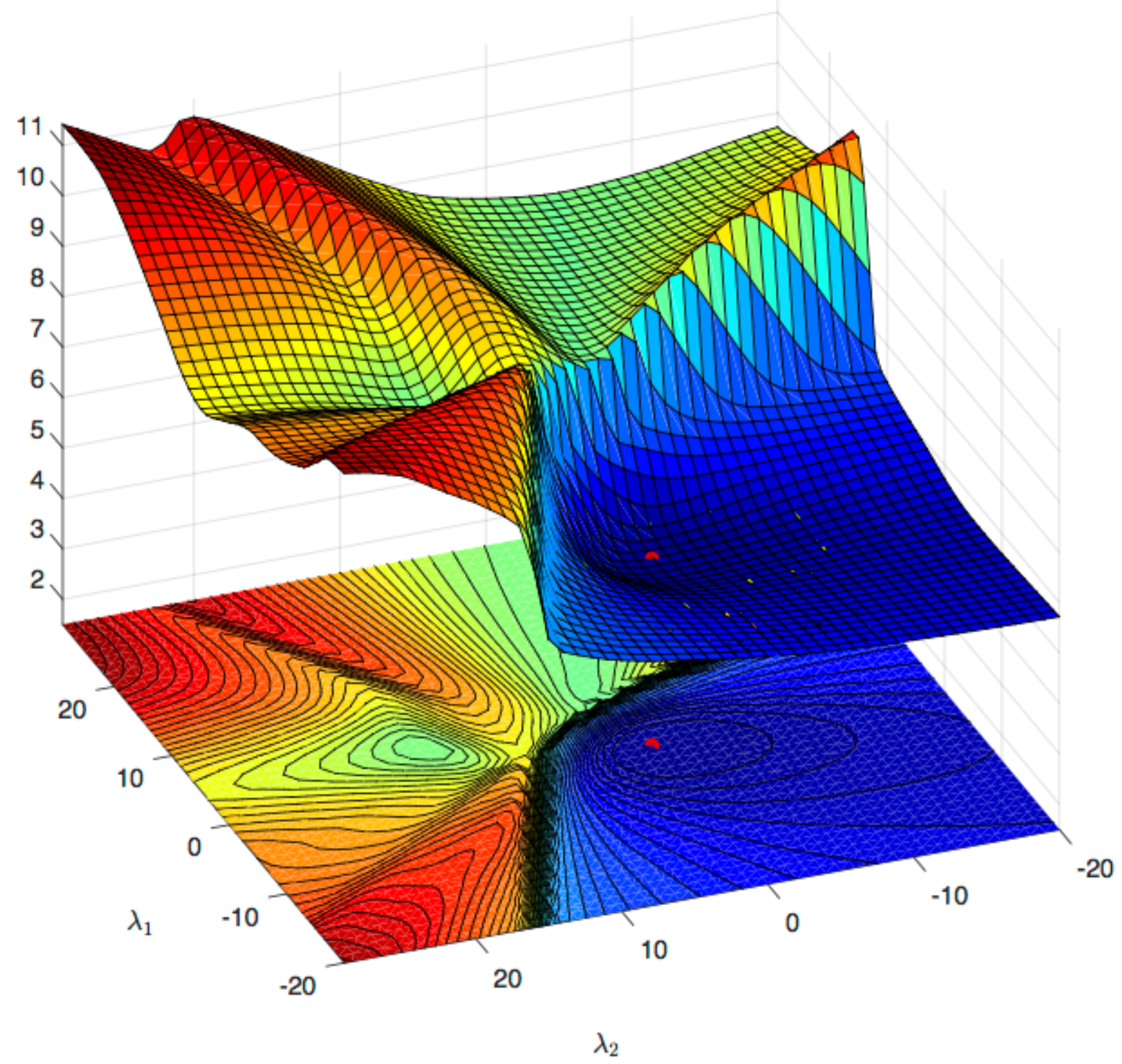}
		\caption{}
	\end{subfigure}	
	\quad
	\begin{subfigure}[t]{0.28\textwidth}
		\includegraphics[width=\textwidth]{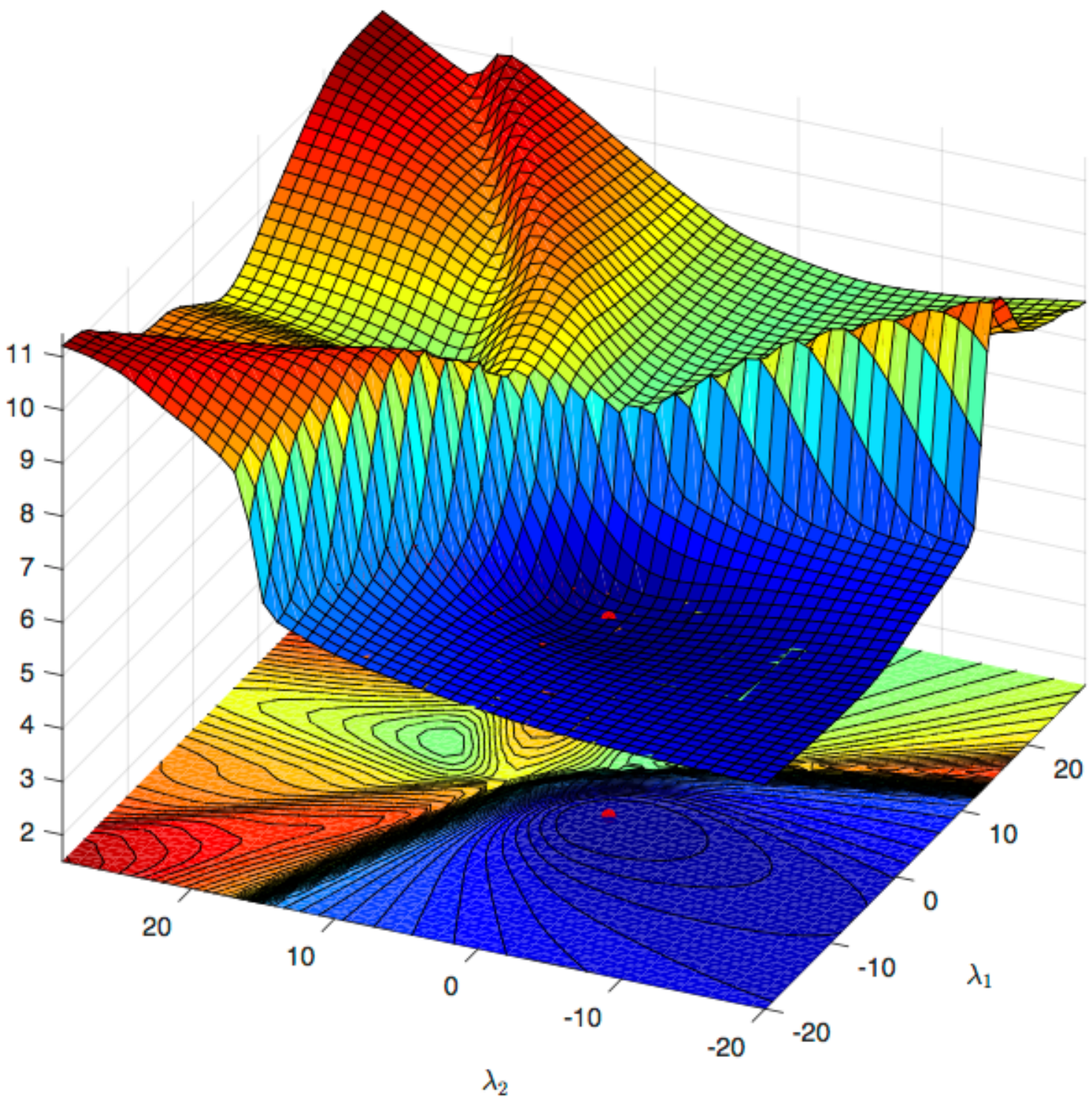}
		\caption{}
	\end{subfigure}
	\quad	
	\begin{subfigure}[t]{0.28\textwidth}
		\includegraphics[width=\textwidth]{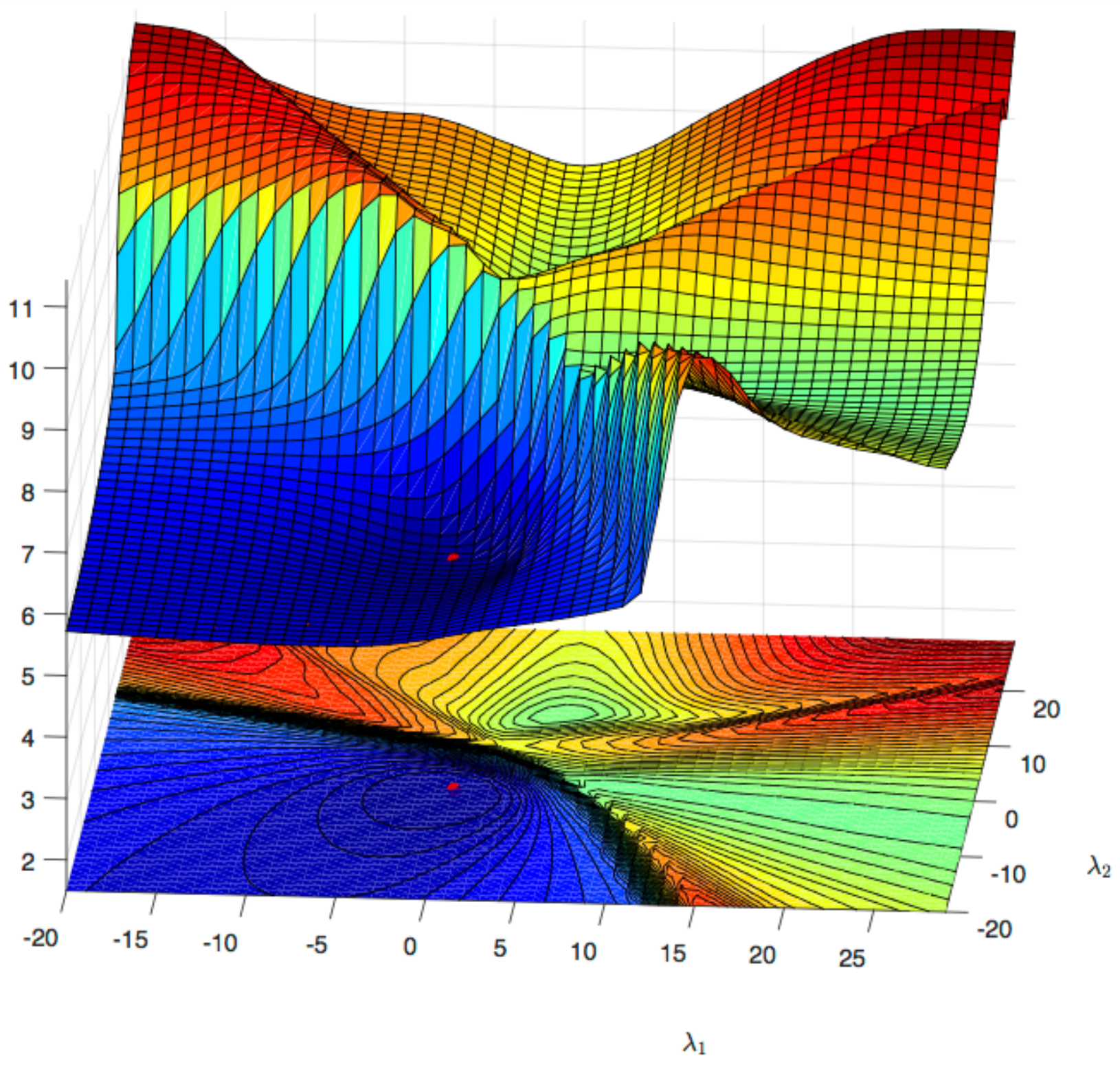}
		\caption{}
	\end{subfigure}	
	\quad
	\begin{subfigure}[t]{0.28\textwidth}
		\includegraphics[width=\textwidth]{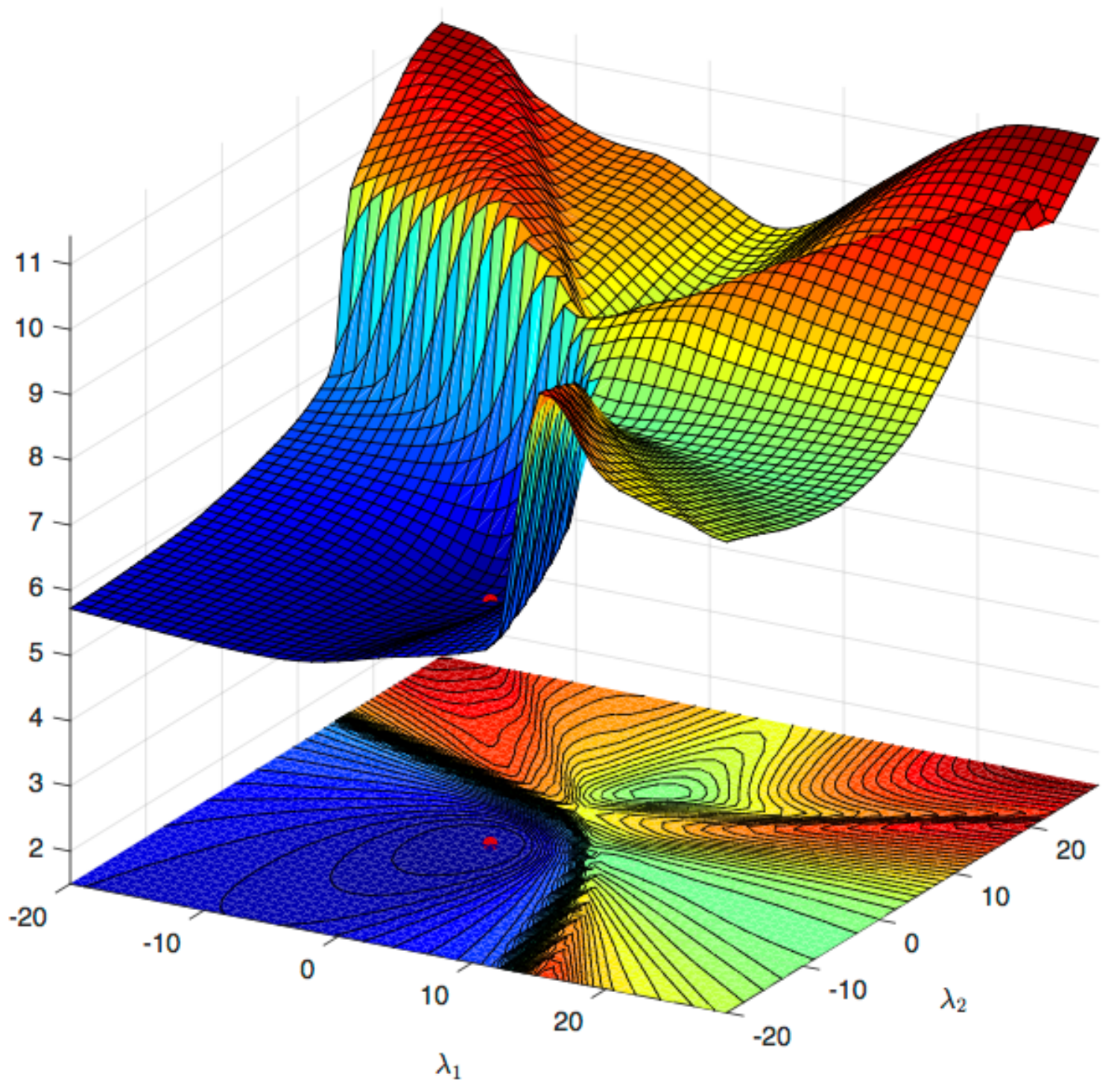}
		\caption{}
	\end{subfigure}	
	\quad
	\begin{subfigure}[t]{0.28\textwidth}
		\includegraphics[width=\textwidth]{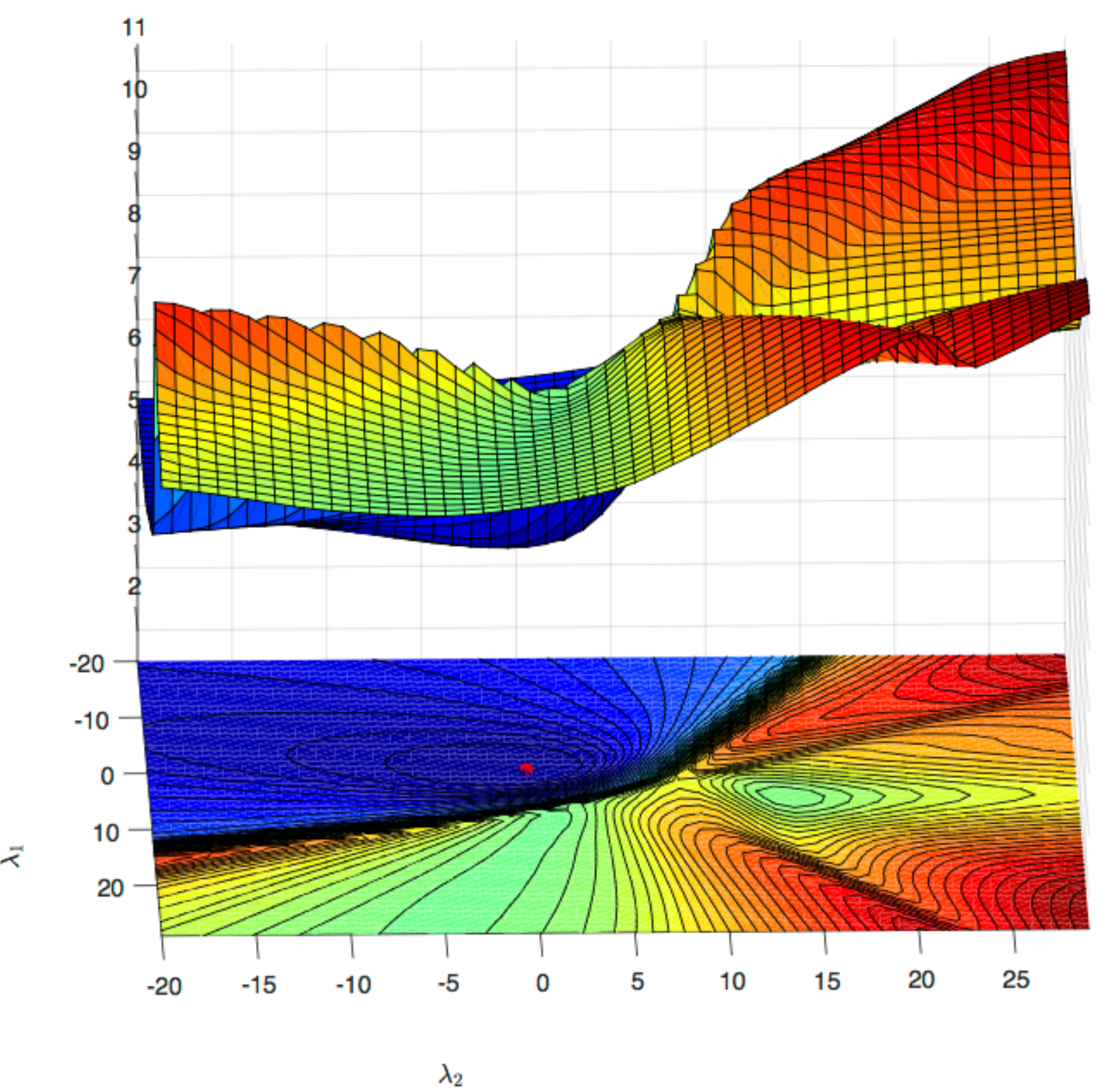}
		\caption{}
	\end{subfigure}
	\quad	
	\begin{subfigure}[t]{0.28\textwidth}
		\includegraphics[width=\textwidth]{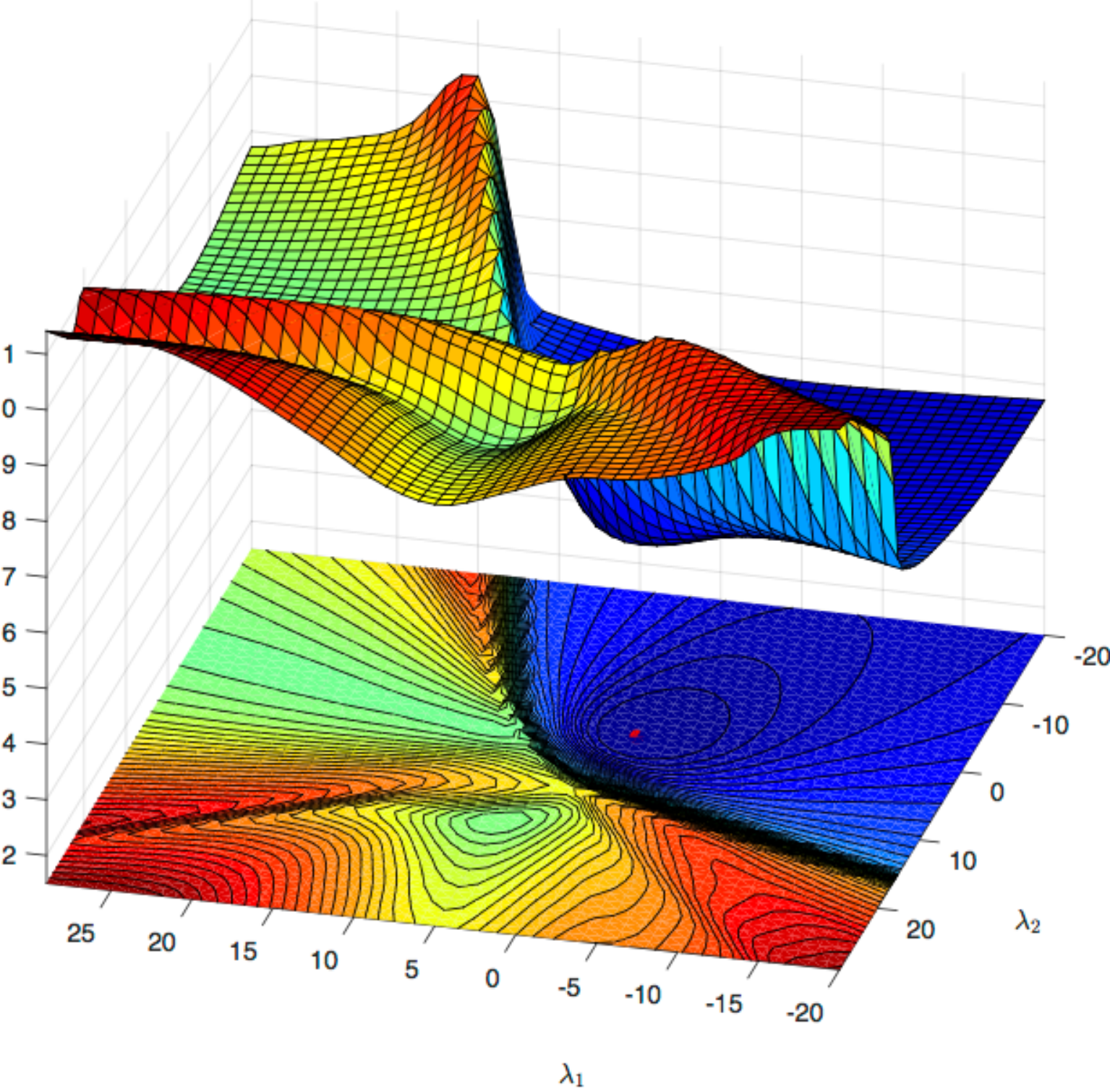}
		\caption{}
	\end{subfigure}	
	\caption{Profile Example II: $\mc{J}(\theta)$ for the OCR dataset on a two-dimensional affine space
	}
	\label{Fig:SurfaceOCR2}
\end{figure*}

\begin{figure*}[t!]
	\centering
	\begin{subfigure}[t]{0.28\textwidth}
		\includegraphics[width=\textwidth]{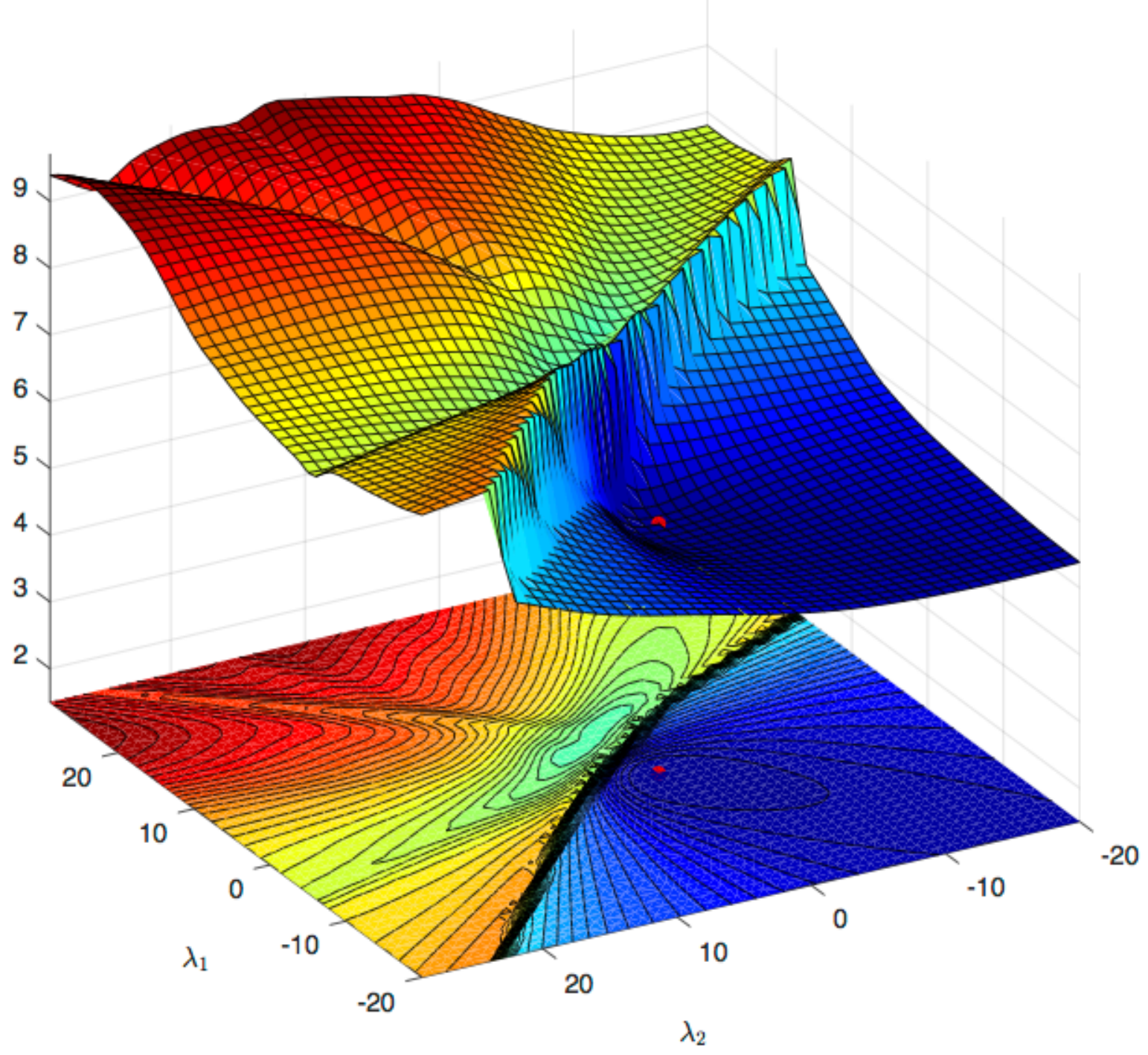}
		\caption{}
	\end{subfigure}	
	\quad
	\begin{subfigure}[t]{0.28\textwidth}
		\includegraphics[width=\textwidth]{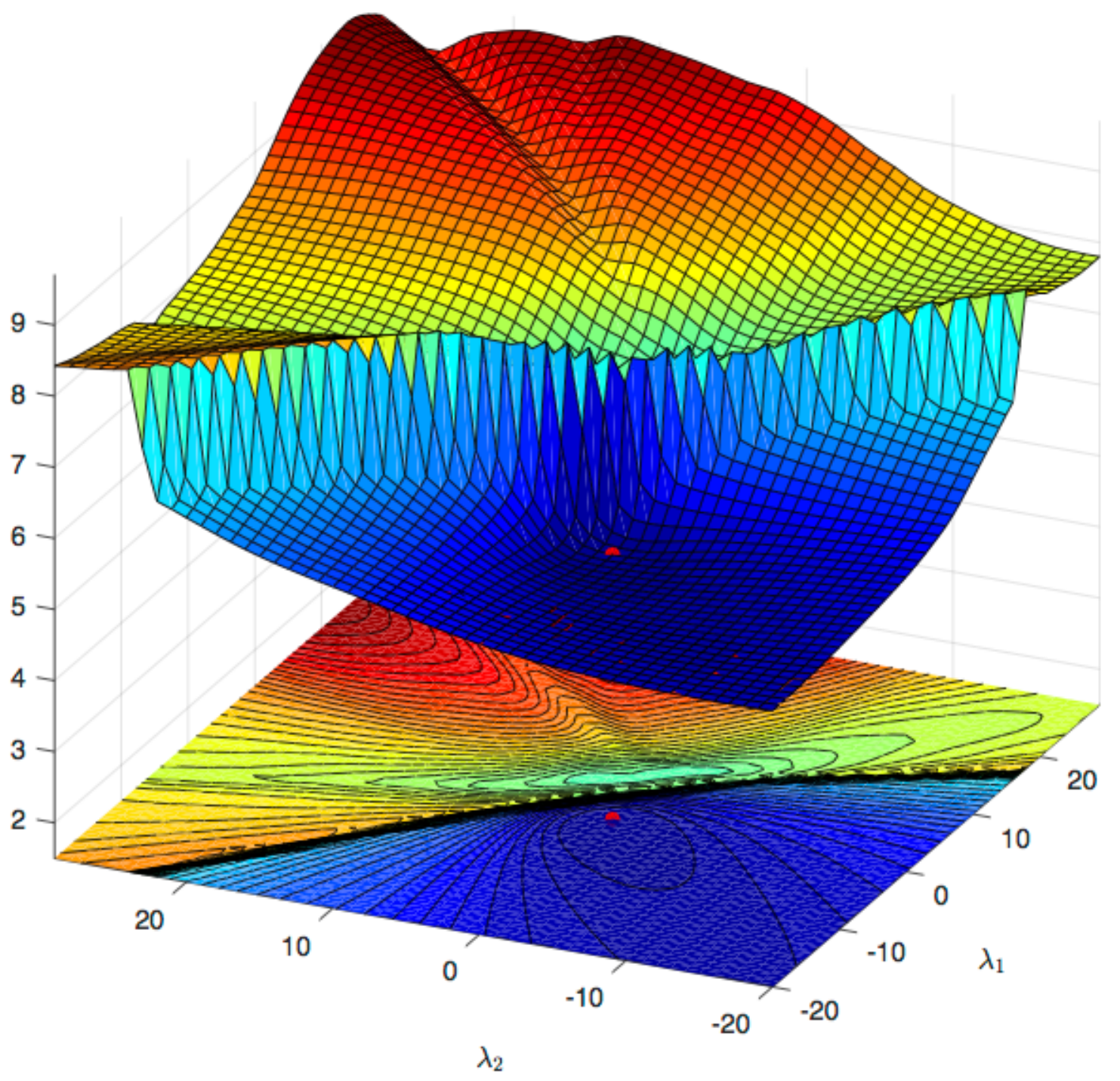}
		\caption{}
	\end{subfigure}
	\quad	
	\begin{subfigure}[t]{0.28\textwidth}
		\includegraphics[width=\textwidth]{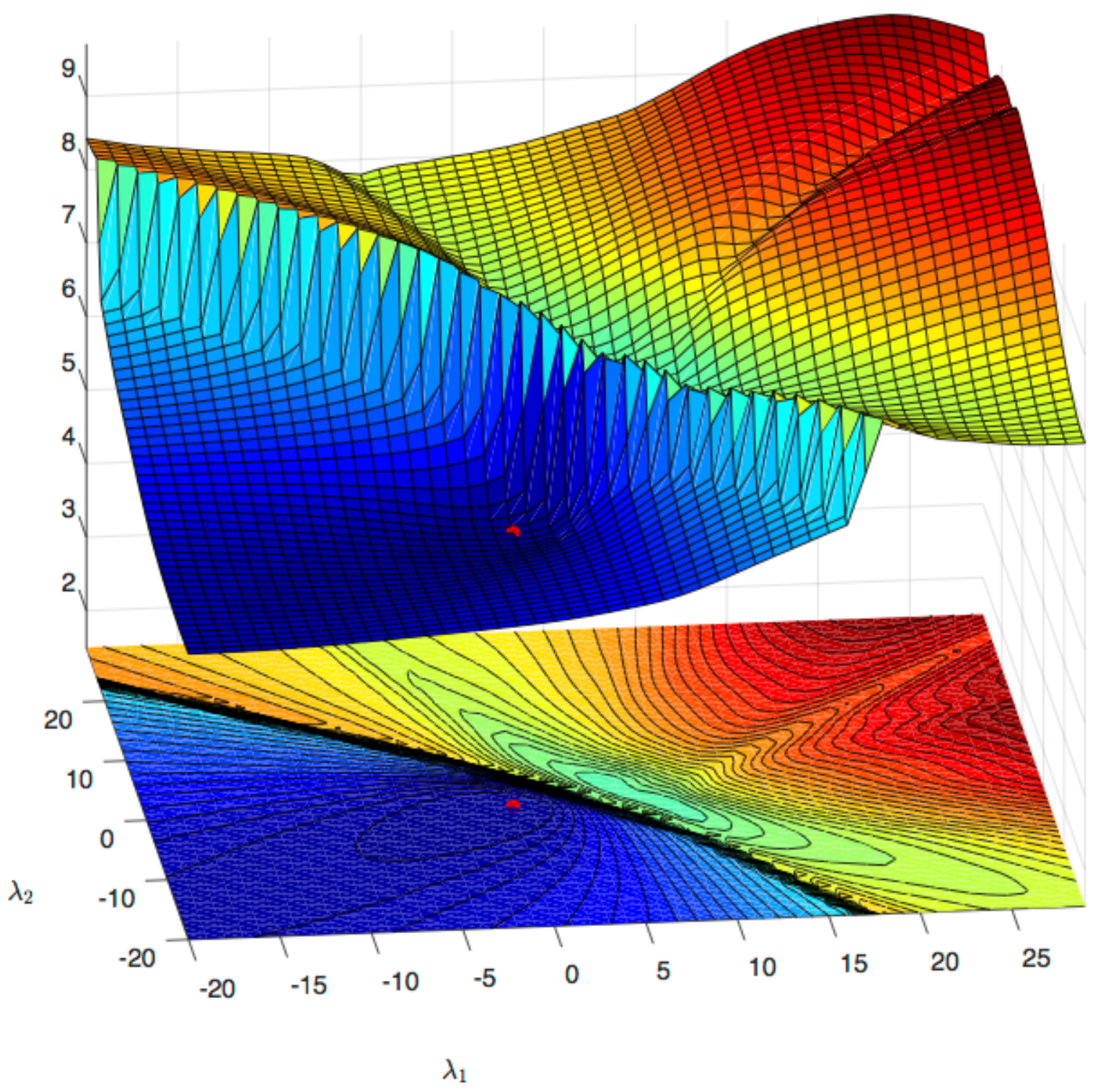}
		\caption{}
	\end{subfigure}	
	\quad
	\begin{subfigure}[t]{0.28\textwidth}
		\includegraphics[width=\textwidth]{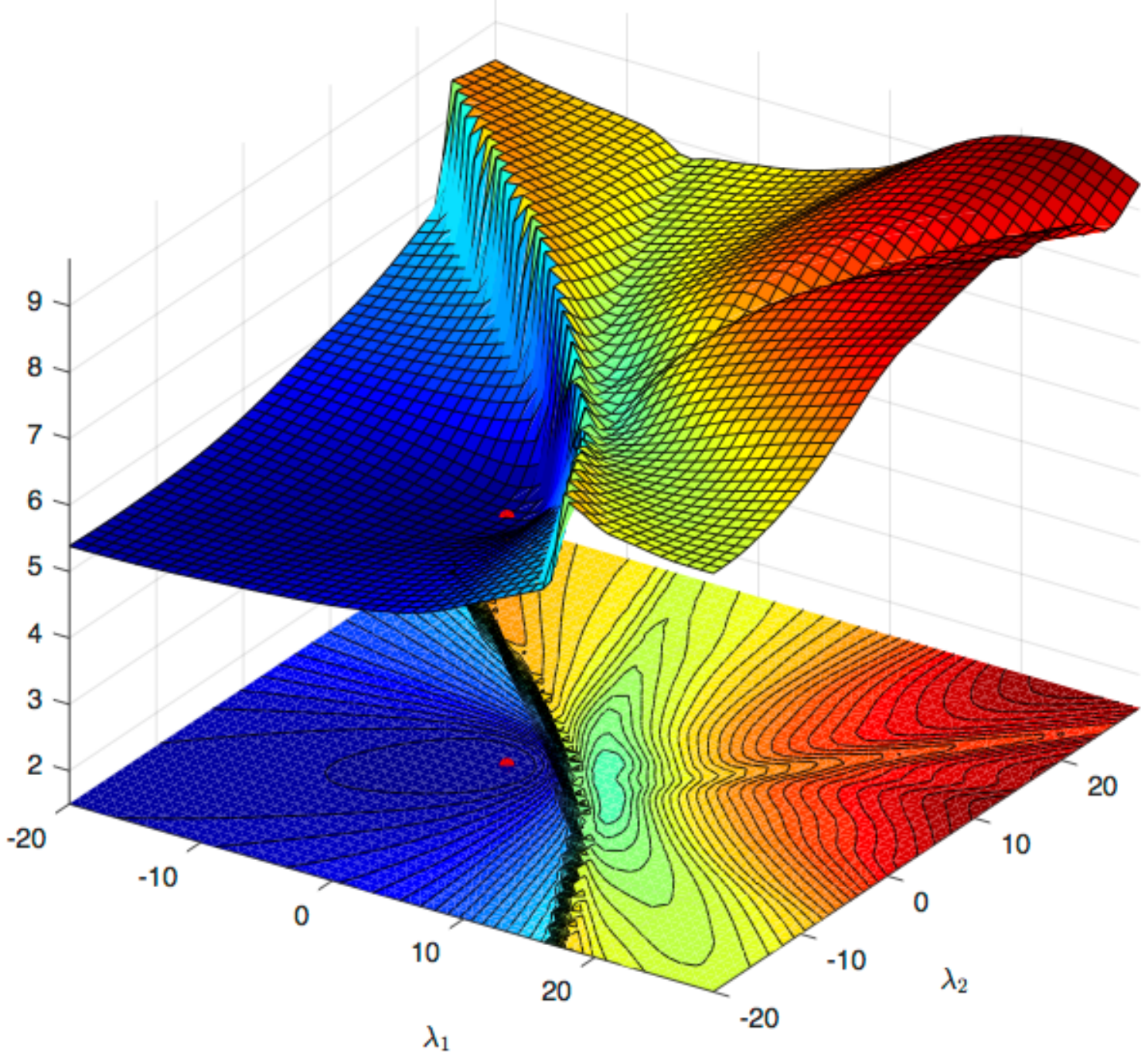}
		\caption{}
	\end{subfigure}	
	\quad
	\begin{subfigure}[t]{0.28\textwidth}
		\includegraphics[width=\textwidth]{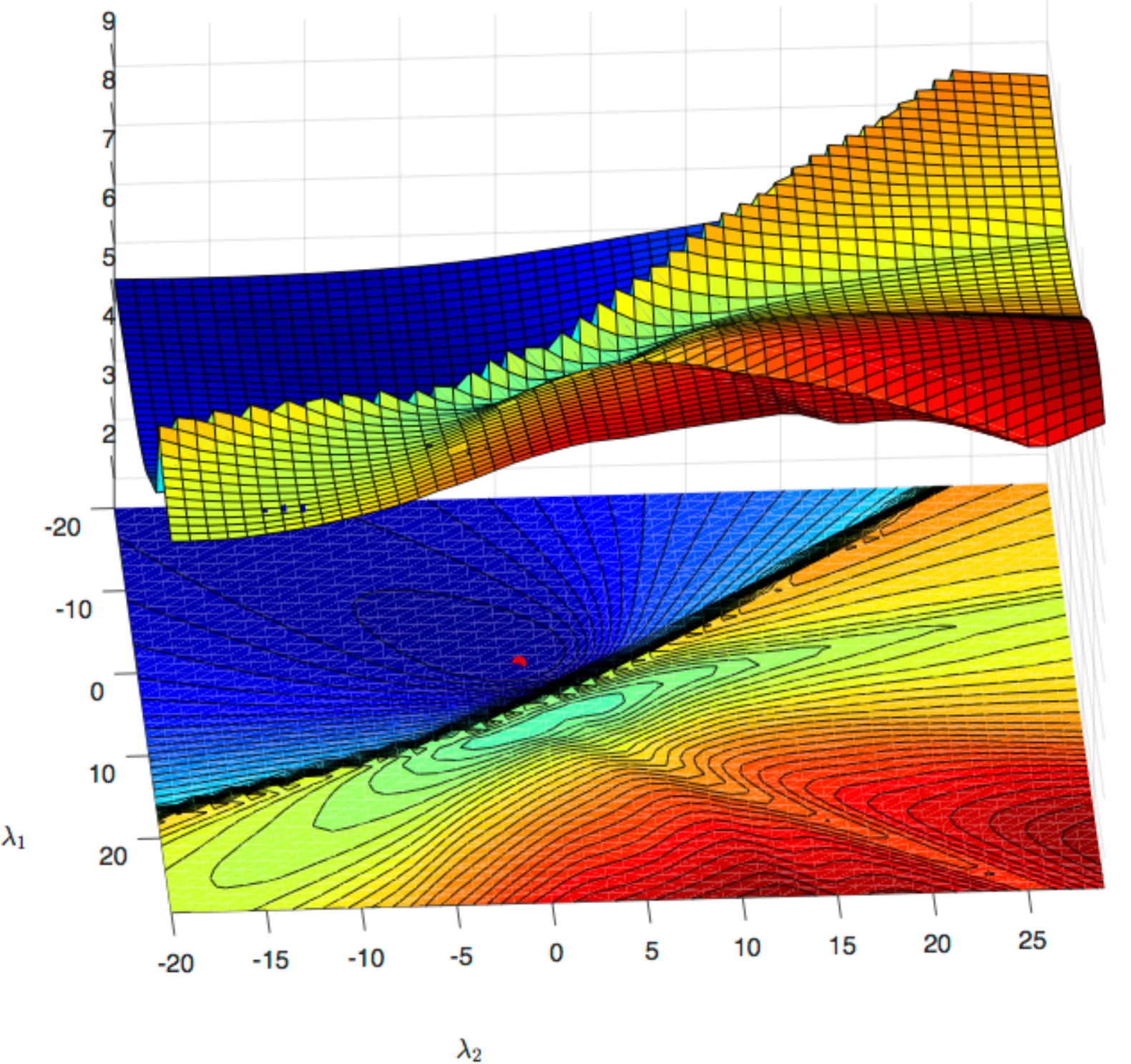}
		\caption{}
	\end{subfigure}
	\quad	
	\begin{subfigure}[t]{0.28\textwidth}
		\includegraphics[width=\textwidth]{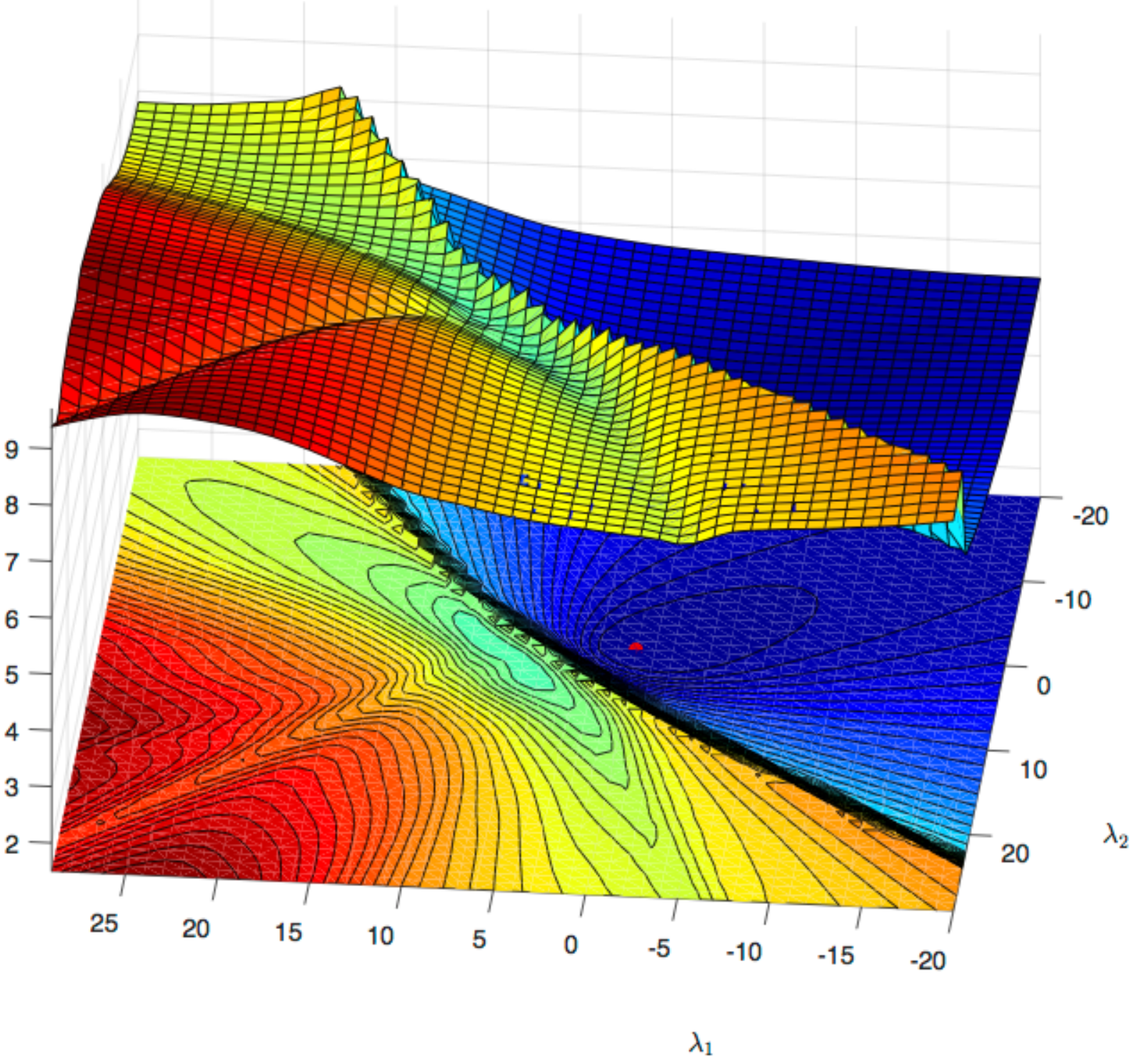}
		\caption{}
	\end{subfigure}	
	\caption{Profile Example III:  $\mc{J}(\theta)$ for the OCR dataset on a two-dimensional affine space
	}
	\label{Fig:SurfaceOCR3}
\end{figure*}

In Figures \ref{Fig:SurfaceOCR1}, \ref{Fig:SurfaceOCR2} and \ref{Fig:SurfaceOCR3}, we show three visualization examples of $\mc{J}(\theta)$ for the OCR dataset on three different affine spaces, part of the first example was included in Figure \ref{Fig:Profile_J_Main}. The six sub-figures in each example show the same profile from six different angles, spinning clock-wise from (a)-(f). The red dots indicate the global minimum.

\begin{figure*}[t!]
	\centering
	\begin{subfigure}[t]{0.28\textwidth}
		\includegraphics[width=\textwidth]{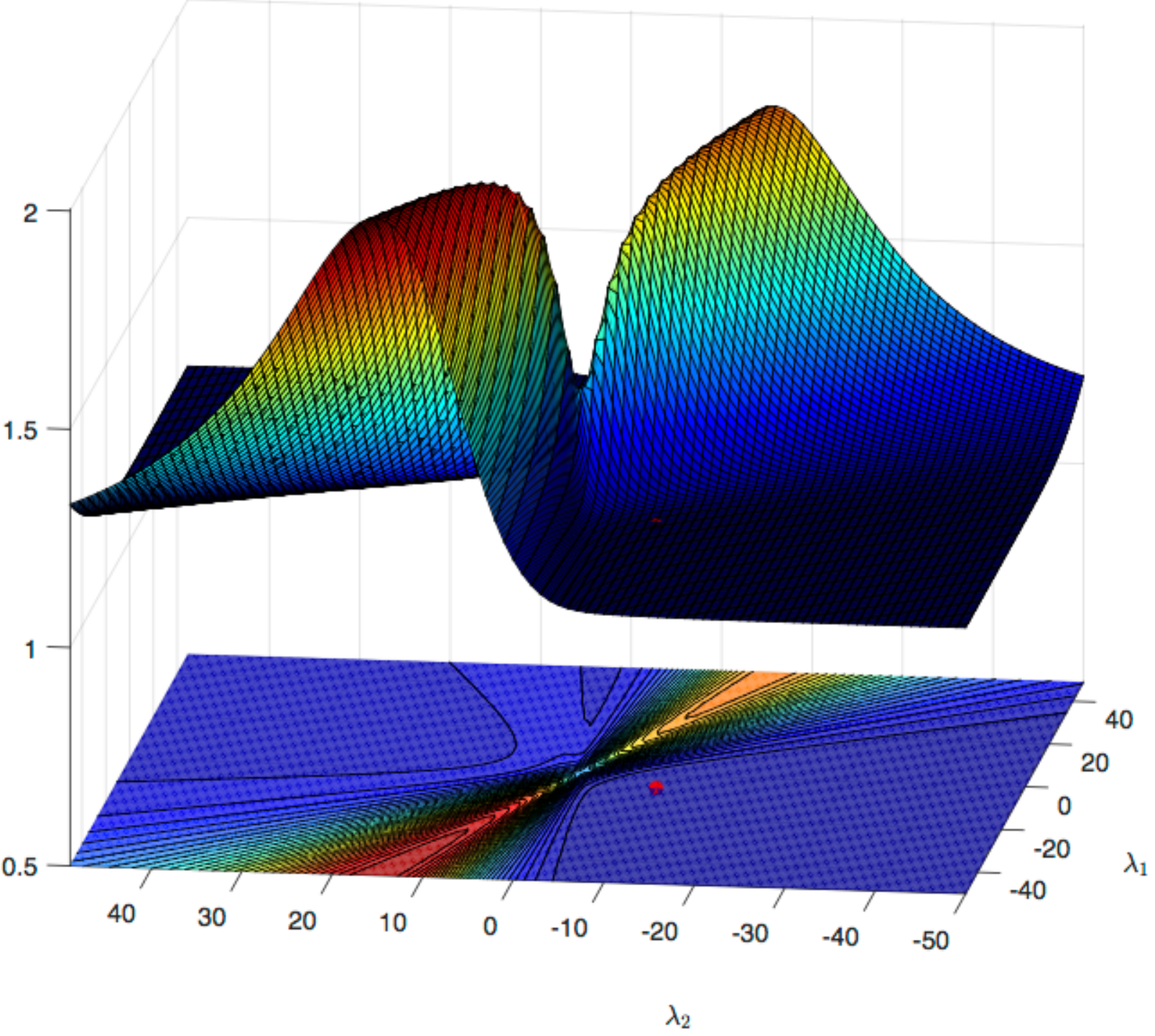}
		\caption{}
	\end{subfigure}	
	\quad
	\begin{subfigure}[t]{0.28\textwidth}
		\includegraphics[width=\textwidth]{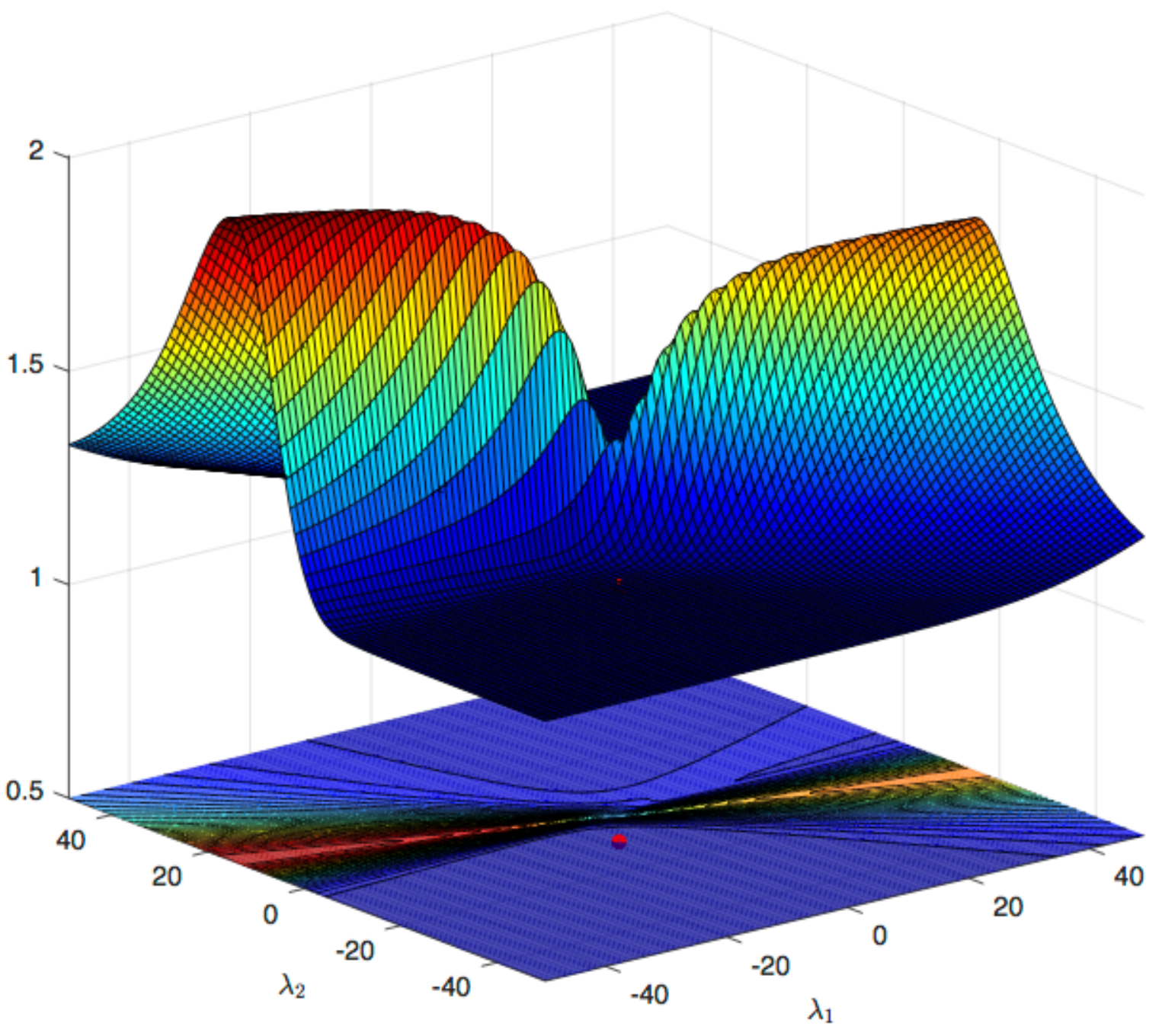}
		\caption{}
	\end{subfigure}
	\quad	
	\begin{subfigure}[t]{0.28\textwidth}
		\includegraphics[width=\textwidth]{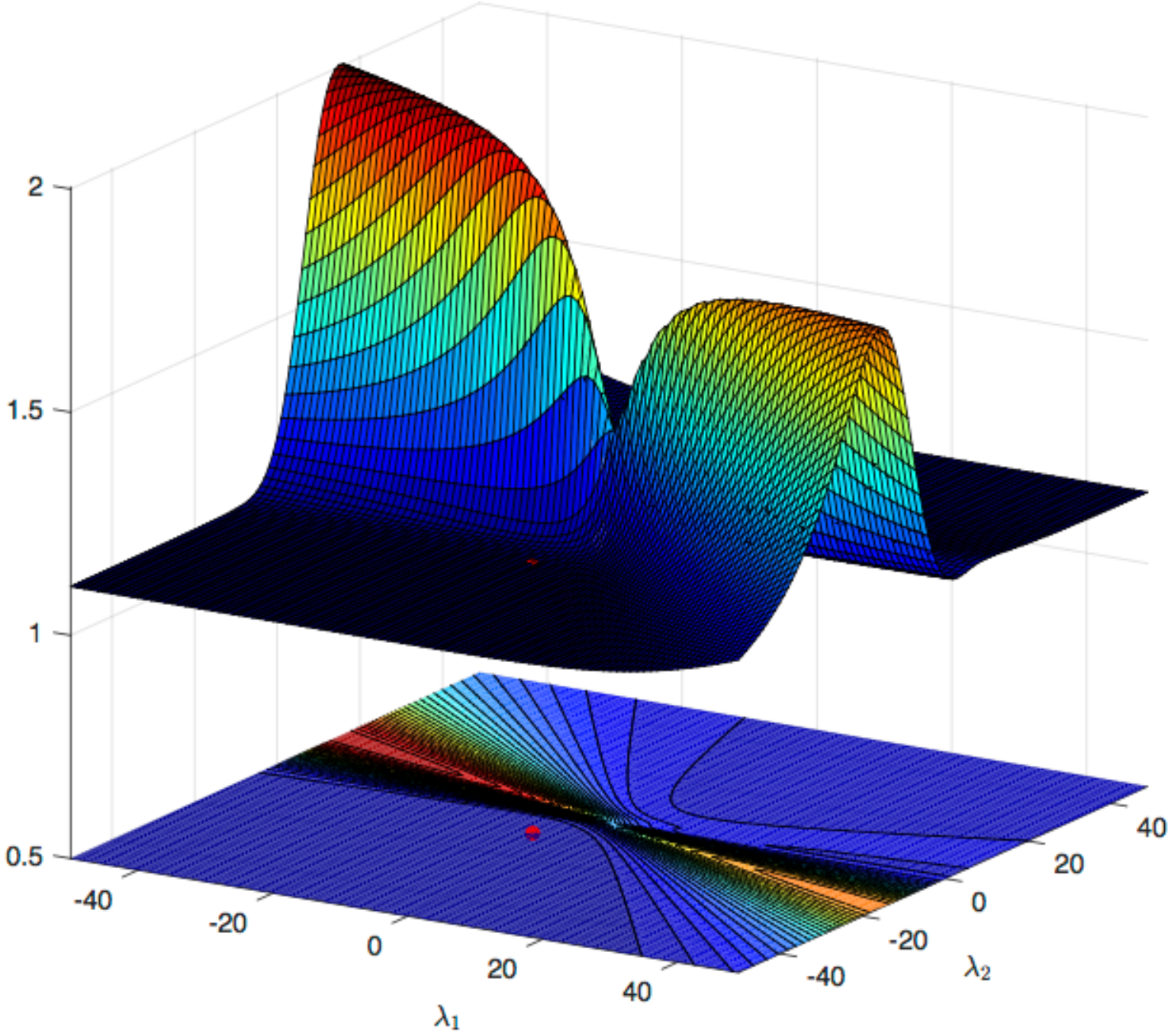}
		\caption{}
	\end{subfigure}	
	\quad
	\begin{subfigure}[t]{0.28\textwidth}
		\includegraphics[width=\textwidth]{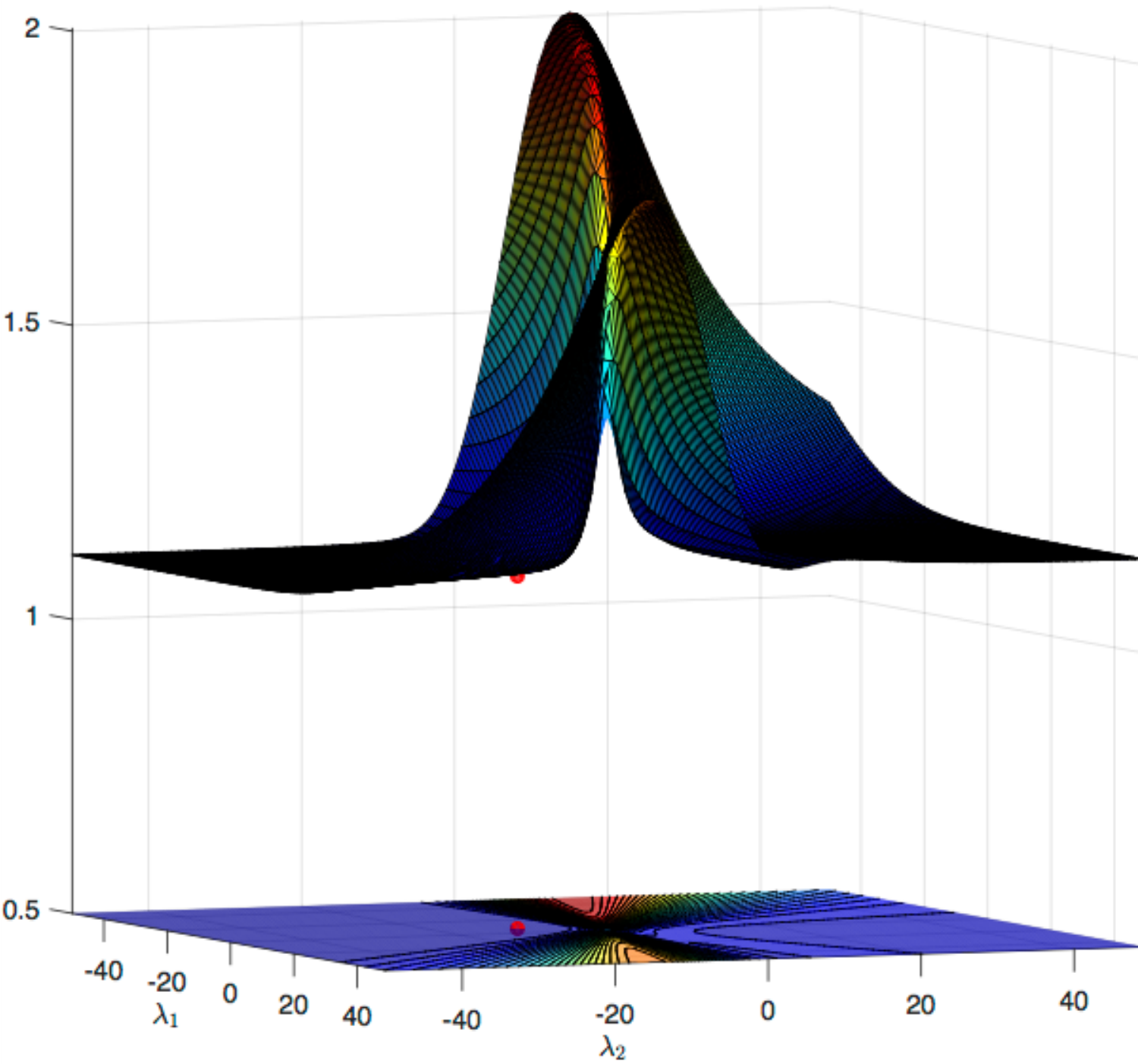}
		\caption{}
	\end{subfigure}	
	\quad
	\begin{subfigure}[t]{0.28\textwidth}
		\includegraphics[width=\textwidth]{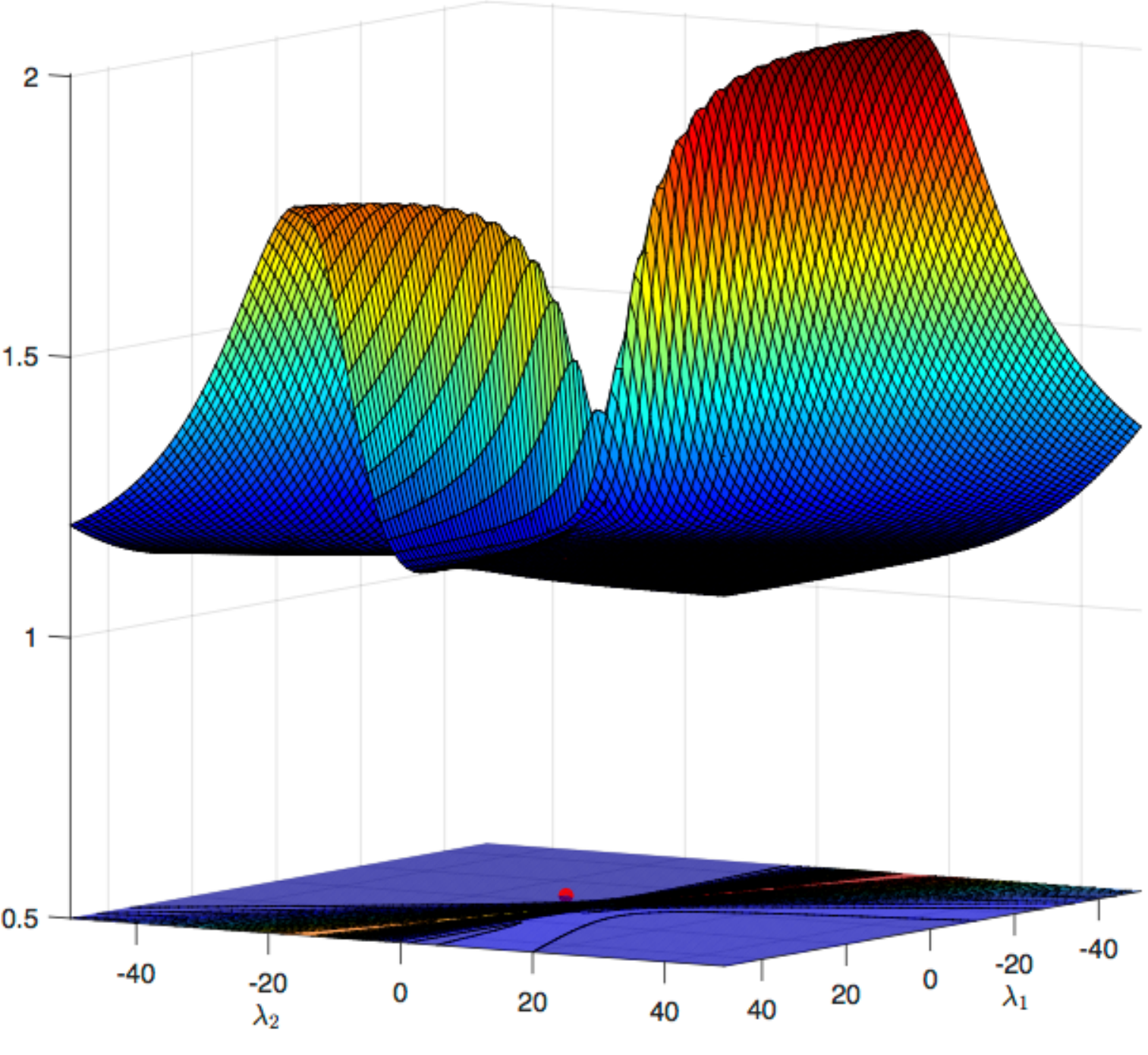}
		\caption{}
	\end{subfigure}
	\quad	
	\begin{subfigure}[t]{0.28\textwidth}
		\includegraphics[width=\textwidth]{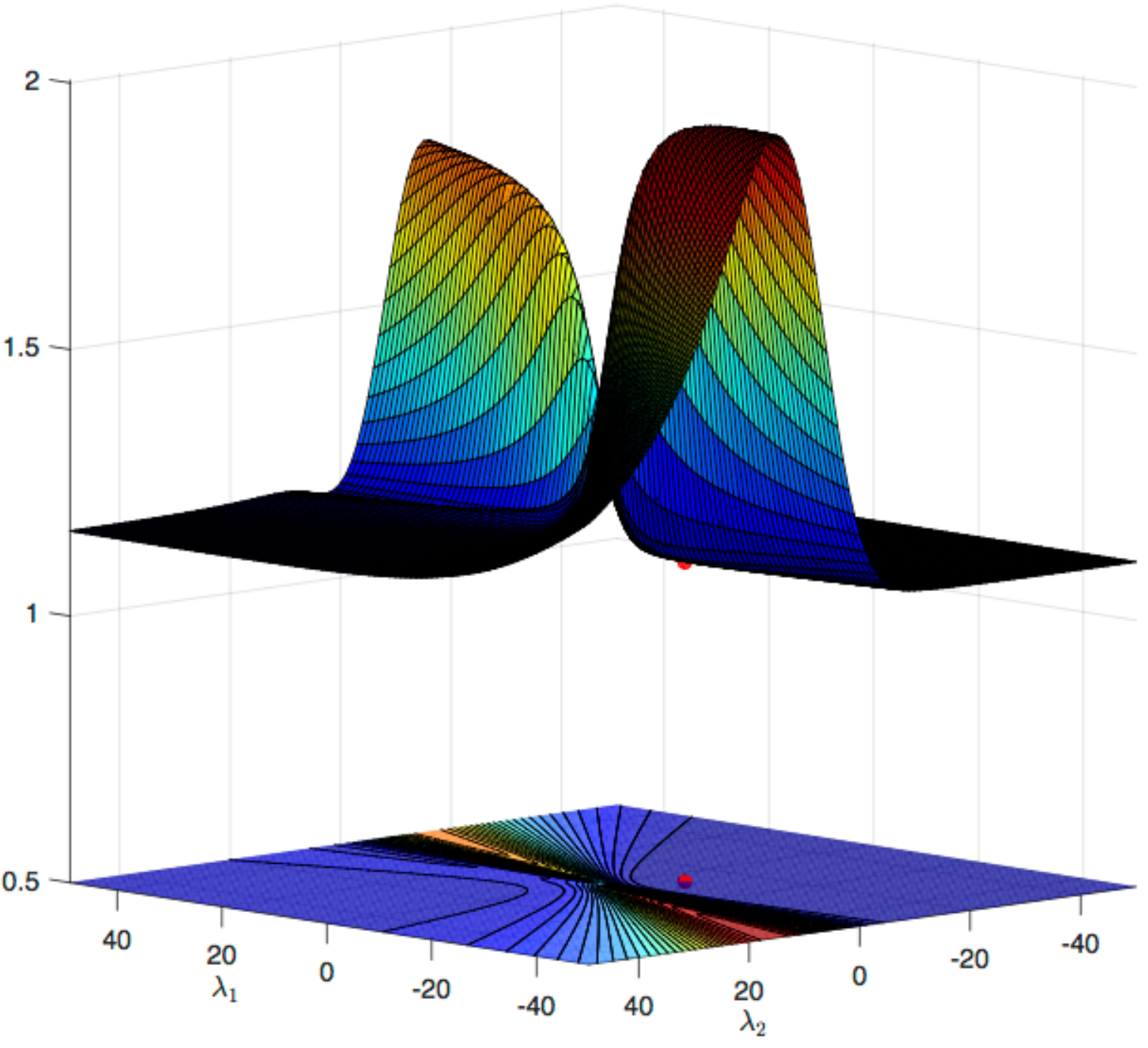}
		\caption{}
	\end{subfigure}	
	\caption{Complete $\mc{J}(\theta)$ profile created from 2-dim synthetic data with two parameters
	}
	\label{Fig:Surface2D}
\end{figure*}

In Figure \ref{Fig:Surface2D}, we show the same type of profiles as above except using synthetic data for of a binary classification problem. First, we sequentially generated a sequence of states from ${0,1}$ by an hidden Markov model. Then we sample the corresponding data points from two separate 2-dimensional Gaussian models.  accordingly.

\section{Additional visualization of $\mc{L}(\theta,V)$}

\begin{figure*}[t!]
	\centering
	\begin{subfigure}[t]{0.28\textwidth}
		\includegraphics[width=\textwidth]{figures/Fig_SaddlePoint_v2}
		\caption{}
	\end{subfigure}	
	\quad
	\begin{subfigure}[t]{0.28\textwidth}
		\includegraphics[width=\textwidth]{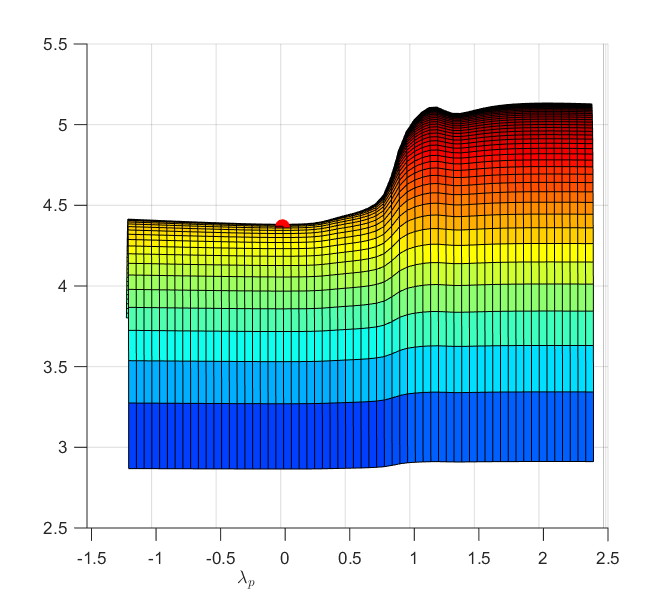}
		\caption{}
	\end{subfigure}
	\quad	
	\begin{subfigure}[t]{0.28\textwidth}
		\includegraphics[width=\textwidth]{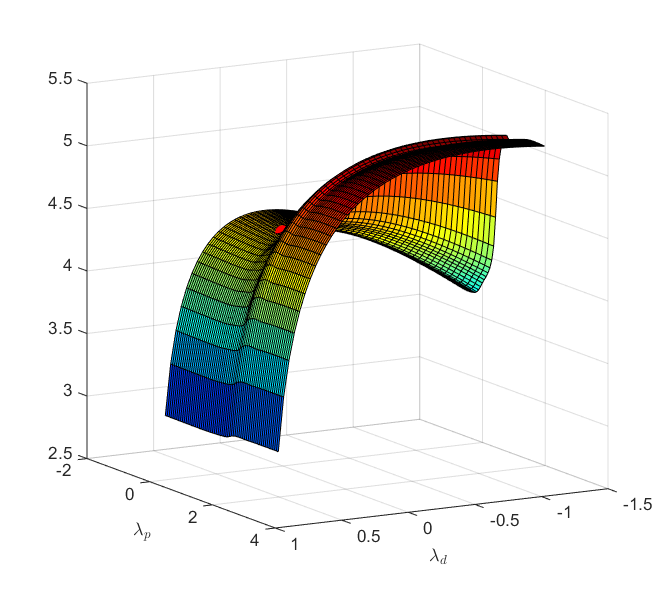}
		\caption{}
	\end{subfigure}	
	\quad
	\begin{subfigure}[t]{0.28\textwidth}
		\includegraphics[width=\textwidth]{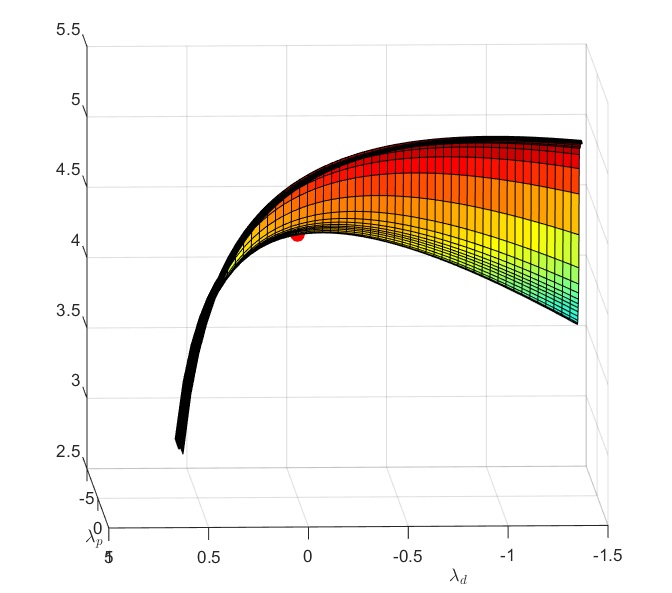}
		\caption{}
	\end{subfigure}	
	\quad
	\begin{subfigure}[t]{0.28\textwidth}
		\includegraphics[width=\textwidth]{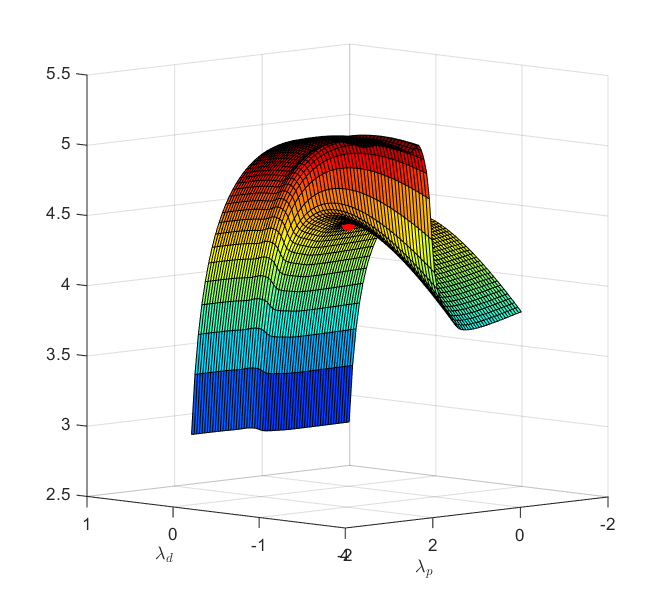}
		\caption{}
	\end{subfigure}
	\quad	
	\begin{subfigure}[t]{0.28\textwidth}
		\includegraphics[width=\textwidth]{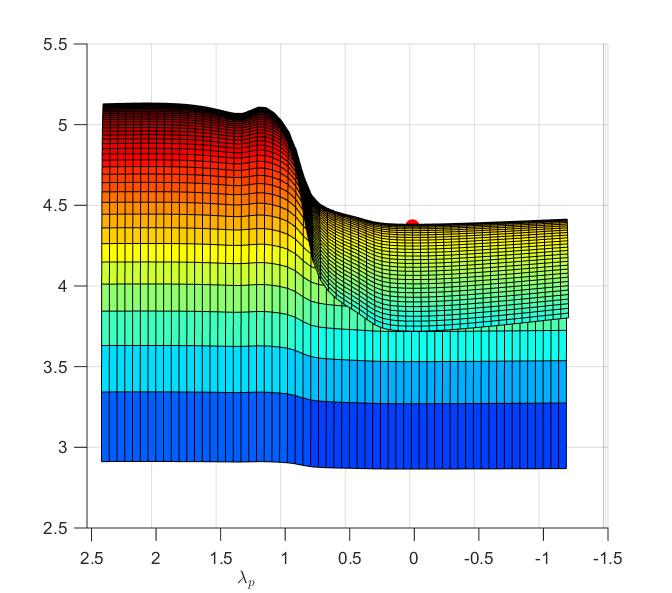}
		\caption{}
	\end{subfigure}	
	\quad
	\begin{subfigure}[t]{0.28\textwidth}
		\includegraphics[width=\textwidth]{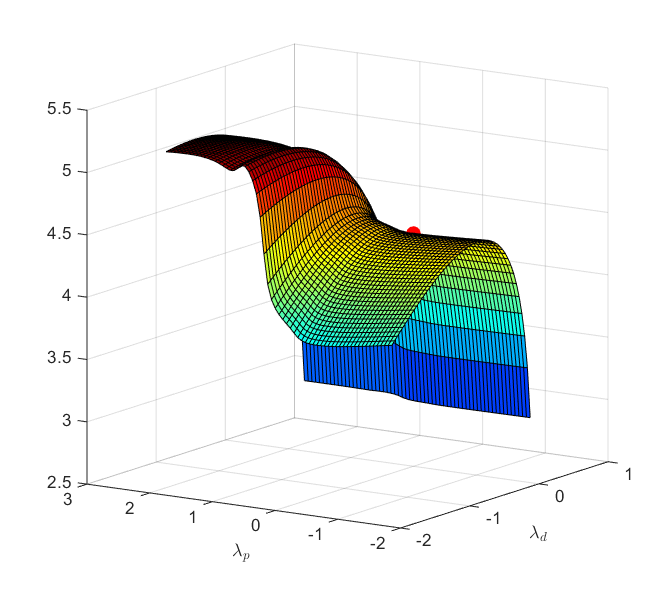}
		\caption{}
	\end{subfigure}	
	\quad
	\begin{subfigure}[t]{0.28\textwidth}
		\includegraphics[width=\textwidth]{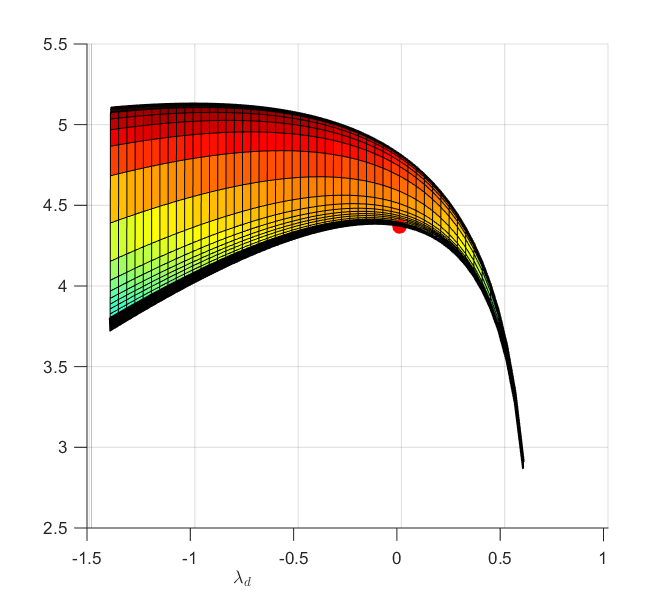}
		\caption{}
	\end{subfigure}
	\quad	
	\begin{subfigure}[t]{0.28\textwidth}
		\includegraphics[width=\textwidth]{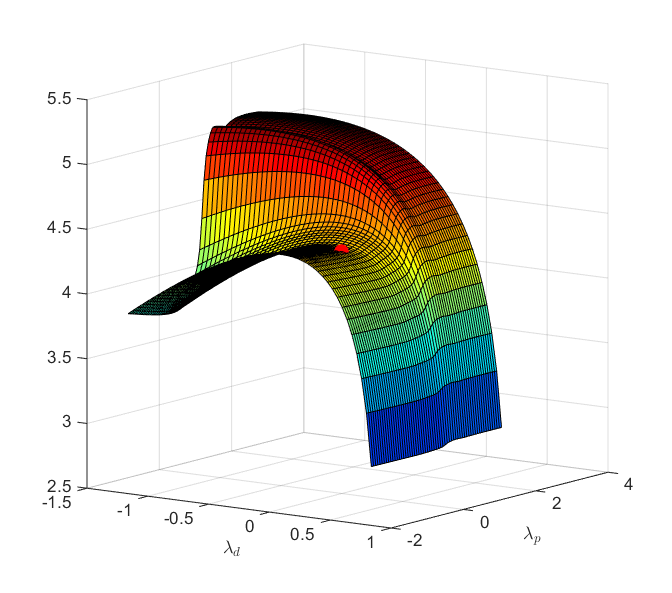}
		\caption{}
	\end{subfigure}	
	\caption{Profile of $\mc{L}(\theta,V)$ for the OCR dataset on a two-dimensional affine space. Red dots show the saddle points (the optimal solution) of the profile from nine different angles.
	}
	\label{Fig:SurfaceSaddle}
\end{figure*}

Figure \ref{Fig:SurfaceSaddle} shows the profile of $\mc{L}(\theta,V)$ for the OCR data set on a two-dimensional affine space viewed from nine different angles. The red dots show the saddle points of the profile, one for each angle.

%\iffalse
% The profiles of $\mc{J}(\theta)$ for the OCR dataset on a two-dimensional affine space that passes through the supervised solution. The three figures show the same profile from different angles, where the red dot is the supervised solution. The contours of the profiles are also shown at the bottom.

%\fi

\end{document}